\documentclass[a4paper,fleqn]{cas-dc}

\usepackage[authoryear]{natbib}

\newcommand{\software}[1]{\texttt{#1}}
\newcommand{\rconv}{\ensuremath{r_{\mathrm{conv}}}}
\newcommand{\rlat}{\ensuremath{r_{\mathrm{lat}}}}

\def\tsc#1{\csdef{#1}{\textsc{\lowercase{#1}}\xspace}}
\tsc{WGM}
\tsc{QE}

\begin{document}
\let\WriteBookmarks\relax
\def\floatpagepagefraction{1}
\def\textpagefraction{.001}

\shorttitle{Latent ODE Modeling of Cine Cardiac MRI}    

\shortauthors{Brüggemann et al.}  

\title [mode = title]{A Latent ODE Approach to Full-Cycle Biventricular Cardiac Motion Modeling from Cine MRI}

\affiliation[1]{organization={Swiss Data Science Center, EPFL and ETH Zürich},
            country={Switzerland}}
\affiliation[2]{organization={Diagnostic and Interventional Radiology, University Hospital Zürich},
            country={Switzerland}}
\affiliation[3]{organization={Institute for Biomedical Engineering, University and ETH Zürich},
            country={Switzerland}}
\affiliation[4]{organization={Biomedical Informatics Group, ETH Zürich},
            country={Switzerland}}
\affiliation[5]{organization={Division of Preventive Medicine, Brigham and Women's Hospital / Harvard Medical School, Boston},
            country={USA}}

\author[1]{D. Brüggemann}[orcid=0000-0002-2409-5548]

\cormark[1]

\ead{david.bruggemann@datascience.ch}

\credit{Conceptualization, Meth\-odology, Software, Formal analysis, Investigation, Writing - Original Draft, Visualization}

\author[1]{E. Krymova}[orcid=0000-0002-5313-3451]

\ead{ekaterina.krymova@datascience.ch}

\credit{Conceptualization, Meth\-odology, Writing - Original Draft, Supervision, Project administration}

\author[1]{F. Ozdemir}[orcid=0000-0001-6643-7318]

\ead{firat.ozdemir@datascience.ch}

\credit{Conceptualization, Meth\-odology, Writing - Original Draft, Visualization}

\author[2,3]{J. {von Spiczak}}[orcid=0000-0002-1978-1535]

\ead{jochen.vonspiczak@usz.ch}

\credit{Conceptualization, Validation, Writing - Original Draft}

\author[3]{S. {Kozerke}}[orcid=0000-0003-3725-8884]

\ead{kozerke@biomed.ee.ethz.ch}

\credit{Meth\-odology, Validation, Writing - Review \& Editing}

\author[5]{S. {Mora}}[orcid=0000-0001-6283-0980]

\ead{smora@bwh.harvard.edu}

\credit{Writing - Review \& Editing, Funding Acquisition}

\author[2]{R. {Manka}}[orcid=0000-0002-3383-4998]

\ead{robert.manka@usz.ch}

\credit{Validation, Writing - Review \& Editing}

\author[1]{M. Salzmann}[orcid=0000-0002-8347-8637]

\ead{mathieu.salzmann@datascience.ch}

\credit{Writing - Review \& Editing, Supervision, Project administration}

\author[4,5]{O. V. Demler}[orcid=0000-0003-3355-3210]

\ead{olga.demler@inf.ethz.ch}

\credit{Conceptualization, Writing - Original Draft, Project administration, Funding acquisition}

\cortext[1]{Corresponding author}

\begin{abstract}
Cardiac magnetic resonance imaging (CMR) captures rich spatiotemporal information about ventricular structure and motion, but conventional risk models use only a few image-derived indices from selected cardiac phases. We present a latent dynamical model that encodes bi-ventricular anatomy and full-cycle cine motion as a continuous latent trajectory, using heart-rate-aware neural ordinary differential equation (ODE) dynamics and a graph-based mesh autoencoder to reconstruct anatomically consistent 3D+t ventricular motion. A covariate-conditioned prior defines the expected end-diastolic latent state, and a Cox proportional hazards model tests whether deviations from this prior predict incident heart failure. We studied 72,386 UK Biobank participants without baseline cardiovascular disease, including 367 incident heart failure events. In a held-out evaluation subset, adding the latent score to refitted pooled cohort equations improved the stratified C-index from 0.704 to 0.785, compared with 0.764 for seven established cardiac markers. Compared with non-graph and non-ODE approaches, the proposed model gave the best trade-off between reconstruction fidelity, generative realism, and downstream prognostic performance. These results suggest that continuous full-cycle modeling of ventricular motion provides informative cardiac phenotypes beyond conventional CMR summaries, while external validation in more representative patient cohorts is required before clinical risk-prediction use.
\end{abstract}


\maketitle

\section{Introduction}\label{sec:introduction}

Cardiovascular disease (CVD) remains the leading cause of death worldwide \citep{who_cvd_global_2025} and heart failure (HF) is a major and growing burden, making early risk stratification essential for timely intervention \citep{meunier2021202,heidenreich20222022}. Cardiac magnetic resonance imaging (CMR) is routinely used for non-invasive assessment of myocardial structure, function, and tissue composition \citep{pennell2010cardiovascular}. Yet risk models recommended by guideline committees use only a few CMR-derived indices, such as left ventricular ejection fraction (LVEF) and myocardial strain, in addition to patient characteristics and biomarkers \citep{meunier2021202,heidenreich20222022}. These metrics are reproducible and interpretable but reduce the data twice: they compress high-dimensional anatomy into hand-crafted summaries and focus on isolated phases, usually end-diastole and end-systole. Subtle abnormalities in the timing, coordination, or trajectory of ventricular contraction and relaxation may therefore be missed despite carrying prognostic signal \citep{bijnens2012myocardial,bello2019deep,qiao2025personalized}. Capturing these abnormalities requires modeling cardiac motion throughout the cycle rather than at isolated phases. Such modeling may help detect early signs of HF across phenotypes ranging from systolic dysfunction with dilation to diastolic dysfunction with impaired relaxation \citep{aziz2013diastolic}.

\begin{figure}
  \vspace{0.5\baselineskip}
  \centering
  \includegraphics[width=\linewidth]{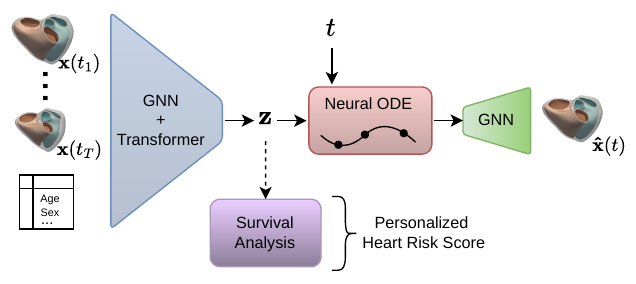}
  \caption{Framework overview. A GNN--Transformer encodes bi-ventricular mesh sequences and subject covariates into a latent vector $\mathbf{z}$. A neural ODE evolves this state continuously, and a graph decoder reconstructs anatomically consistent cardiac motion $\hat{\mathbf{x}}(t)$. Survival analysis links $\mathbf{z}$ to future outcomes to produce a personalized heart failure risk score.}\label{fig:teaser}
\end{figure}

Large-scale datasets such as the UK Biobank \citep{petersen2016uk,littlejohns2020ukbiobank} now provide tens of thousands of CMR scans with longitudinal follow-up. They create an opportunity to study how cardiac shape and motion relate to future risk, but require models that are expressive, physiologically plausible, and scalable.

Existing models only partly meet this need. Generative approaches have captured population-level anatomical variability \citep{beetz2022interpretable,dou2023conditional}, but largely ignore motion. Recent spatiotemporal models track mesh motion or learn frame-wise latent trajectories \citep{meng2024deepmesh,qiao2025personalized}, but treat time as a normalized discrete sequence. This is limiting for cine CMR, where frame timing varies with heart rate and systole and diastole occupy different fractions of the cycle across subjects.

We therefore model bi-ventricular motion as a continuous latent trajectory (see Figure~\ref{fig:teaser}). The model starts from vertex-registered ventricular meshes, which provide an anatomically consistent representation of cardiac shape \citep{frangi2003automatic}. Graph neural networks preserve the mesh connectivity \citep{bronstein2017geometric}, while neural ODEs evolve the latent state smoothly across arbitrary cardiac phases \citep{chen2018neural}. We embed this dynamical model in a covariate-conditioned variational autoencoder (VAE) \citep{kingma2013auto,sohn2015learning}: age, sex, and body surface area (BSA) define the expected end-diastolic latent state, the observed mesh sequence defines a subject-specific posterior, and the standardized difference between the two yields residual spatiotemporal embeddings for Cox survival analysis.

In summary, this work contributes a continuous latent ODE model for full-cycle CMR shape and motion, a covariate-conditioned prior for physiologically structured representations, heart rate-aware phase warping for timing variability, and a UK Biobank evaluation showing that latent spatiotemporal deviations add prognostic information beyond standard clinical and CMR indices.

\section{Related Work}\label{sec:related_work}

\paragraph{Mesh-based cardiac shape and motion analysis.}
Statistical shape models provide a principled description of ventricular anatomy \citep{frangi2003automatic,fonseca2011cardiac}, while benchmark studies have established shape-based LV classification as a quantitative standard \citep{suinesiaputra2017statistical}. Mesh representations extend this idea to 3D motion over the cardiac cycle, enabling spatiotemporal descriptors of infarction \citep{piras2017morphologically} and biventricular atlases of shape and motion \citep{bai2015bi}. At population scale, UK Biobank studies show that atlas-derived shape components relate strongly to cardiovascular risk factors and support automatic 3D+t four-chamber quantification \citep{gilbert2019independent,mauger2019right,xia2022automatic}. Mesh-based remodeling signatures also improve long-term HF/CVD event prediction and post-infarction risk stratification beyond conventional CMR measures \citep{mauger2022multi,corral2022understanding}. More recently, template-deformation and graph-convolutional networks have enabled accurate image-to-mesh reconstruction from cardiac CMR \citep{kong2021wholeheart,kong2023learning,gaggion2025multi,mercadier2025printmesh}. Together, these studies show that mesh-based features capture clinically relevant remodeling and can be extracted at scale. Most, however, treat shape and motion as fixed descriptors rather than as samples from a probabilistic model of full-cycle dynamics.

\paragraph{Generative and spatiotemporal cardiac models.}
Early generative models focused on static anatomy. Conditional flow-based VAEs, graph VAEs, and 3D convolutional generative models have been used to synthesize LV anatomies, capture mesh-based shape variability, and learn interpretable remodeling patterns \citep{dou2023conditional,beetz2022interpretable,biffi2018learning}. These models characterize anatomical variation but do not model motion through the cardiac cycle. Recent spatiotemporal approaches address this gap by tracking mesh motion across cine CMR \citep{meng2024deepmesh} or by combining graph encoders with temporal transformers or recurrent modules to learn latent motion trajectories \citep{qiao2025personalized,dou20254d}. These methods still represent time as a discrete frame sequence, implicitly assuming uniform sampling across subjects. Cine CMR violates this assumption: frame counts and/or temporal spacing can vary with heart rate, and systole and diastole occupy different fractions of the cycle.

We instead model dynamics with a continuous-time latent ODE \citep{chen2018neural}. A learned vector field evolves a single initial condition into a smooth latent trajectory defined at arbitrary phases, making the representation compatible with variable frame timing and physiological phase normalization. This continuous-flow view is related to diffeomorphic registration, where velocity fields generate smooth deformations over time \citep{beg2005computing}. Probabilistic and learning-based variants have been applied to cine CMR and large-scale deformable registration \citep{krebs2019learning, dalca2019unsupervised}, and neural ODEs have been used to model registration as particle trajectories \citep{wu2022nodeo}. These methods evolve image- or coordinate-space deformations for registration. Our model instead learns generative cardiac dynamics in a low-dimensional latent space of mesh geometry, with the initial condition summarizing subject-specific motion over the full cycle.

\paragraph{Motion-based cardiac risk stratification.}
Cardiac motion features have also been linked to clinical risk. Survival models on 3D right-ventricular motion fields predict outcomes in pulmonary hypertension \citep{dawes2017machine}, and dense 3D motion fields encoded with a Cox-supervised denoising autoencoder yield mortality-predictive representations \citep{bello2019deep}. For heart failure, motion representations based on myocardial border extraction and dense optical flow have been coupled to multi-modal deep Cox models \citep{guo2023survival}. Our approach differs by first learning a covariate-conditioned normative model of full-cycle motion, then deriving a Cox risk score from latent deviations from this norm.

\section{Methods}\label{sec:methods}

\begin{figure*}
  \centering
  \includegraphics[width=\linewidth]{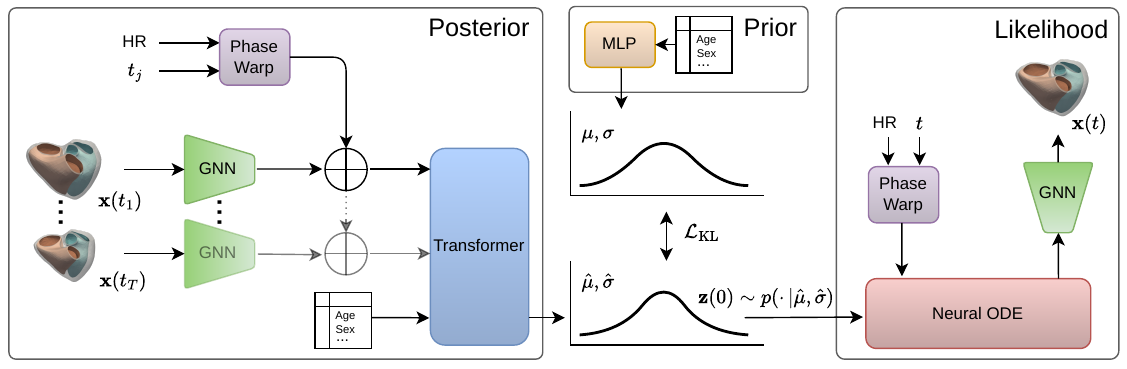}
  \caption{Probabilistic framework for bi-ventricular 3D+t cardiac motion. Left: Frame-wise graph neural networks (GNNs) encode a mesh sequence $\mathbf{x}(t_1),\ldots,\mathbf{x}(t_T)$; heart-rate-aware phase embeddings and a temporal transformer infer an approximate posterior over the end-diastolic latent state $\mathbf{z}(0)$. Middle: An MLP predicts a covariate-conditioned prior on $\mathbf{z}(0)$, and KL regularization aligns posterior and prior while preserving subject-specific deviations. Right: A neural ordinary differential equation (ODE) evolves the sampled latent state over warped cardiac phase, and a graph decoder reconstructs ventricular meshes $\mathbf{x}(t)$ at arbitrary times. HR: heart rate, $\mathcal{L}_{\mathrm{KL}}$: KL divergence loss.}\label{fig:schematic}
\end{figure*}

We model bi-ventricular cardiac motion from 3D+t meshes\footnote{A mesh is a 3D surface representation consisting of triangular faces, defined by vertices (points in 3D space) and edges connecting them.} with fixed topology and vertex correspondence across subjects and frames (with $V$ vertices per mesh). Let $\mathbf{x}(t) \in \mathbb{R}^{3V}$ denote the stacked vertex coordinates at time $t$, where $t \in [0,1]$ denotes a normalized, continuous, within-cycle time variable. We fix $t=0$ at end-diastole (ED) and $t=1$ at the next ED.

As summarized in Figure~\ref{fig:schematic}, a graph encoder infers an approximate posterior over a subject-specific end-diastolic latent state, $\mathbf{z}(0)$, from the observed mesh sequence. A heart-rate-aware neural ODE evolves this state through the cardiac cycle, and a graph decoder maps the latent trajectory back to meshes at any phase. In parallel, a covariate-conditioned prior specifies the expected $\mathbf{z}(0)$ given age, sex, and BSA. Variational training then encourages the posterior to capture subject-specific deviations beyond normal physiological variation, yielding a compact representation for reconstruction and risk modeling.

\subsection{Covariate-Conditioned Latent Cardiac Representation}

Each subject is represented by a continuous latent trajectory $\mathbf{z}(t) \in \mathbb{R}^{D}$. 
We place a covariate-conditioned Gaussian prior on the latent state at end-diastole ($t=0$):
\begin{equation}
\mathbf{z}(0) \sim p\bigl(\mathbf{z}(0)\mid \mathbf{u}\bigr)
= \mathcal{N}\!\left(\boldsymbol{\mu}(\mathbf{u}),\,\mathrm{diag}\!\left(\boldsymbol{\sigma}^2(\mathbf{u})\right)\right),
\label{eq:cond_prior}
\end{equation}
where $\mathbf{u}$ contains age, sex, and BSA. A shared one-hidden-layer MLP (width 64, ReLU) with separate linear heads parameterizes $\boldsymbol{\mu}(\mathbf{u})$ and $\boldsymbol{\sigma}(\mathbf{u})$.
Conditioning on $\mathbf{u}$ organizes the latent space by known physiological variation and encourages the model to isolate subject-specific deviations from population trends, consistent with the normative modeling paradigm \citep{marquand2016understanding}. Furthermore, it establishes a link to the identifiability literature (see Section~\ref{subsec:var}).

\subsection{Continuous Latent ODE Dynamics}
\label{subsec:latent_dynamics}

Starting from $\mathbf{z}(0)$, the latent state at end-diastole (a phase revisited at $t=1$), we model temporal evolution using a latent ordinary differential equation (ODE) \citep{chen2018neural}:
\begin{equation}
\frac{d\mathbf{z}(t)}{dt}
= \mathbf{f}\bigl(\mathbf{z}(t),\sin(\theta(t)), \cos(\theta(t))\bigr),
\label{eq:latent_ode}
\end{equation}
$\mathbf{f}(\cdot)$ is a neural vector field and $\theta(t) \in [0,2\pi]$ is a latent phase variable (Section~\ref{subsec:phase_warp}). The circular embedding $[\sin(\theta),\cos(\theta)]$ adds a periodic bias while allowing flexible, non-sinusoidal motion.
This formulation does not explicitly enforce $\mathbf{z}(0) = \mathbf{z}(1)$. A cycle-consistency penalty did not improve modeling fidelity or downstream clinical performance in preliminary experiments, suggesting that quasi-periodicity is learned implicitly; we therefore omit it.

The vector field $\mathbf{f}$ is a three-layer MLP (hidden size 64, $\tanh$ activations). Unlike standard discrete-time decoders, including transformers and recurrent neural networks, our latent ODE yields a trajectory that is smooth by construction through integration of Eq.~\eqref{eq:latent_ode} over $t \in [0,1]$. The trajectory can be evaluated at arbitrary times, natively accommodating irregular frame timing and variable frame counts without resampling observations onto a common temporal grid. The shared $\mathbf{f}$ captures common cardiac dynamics; the initial state $\mathbf{z}(0)$ determines the subject-specific trajectory across the phase $\theta(t)$ and provides a fixed-dimensional representation for downstream risk analysis.

\subsection{Heart Rate-Aware Phase Warping}
\label{subsec:phase_warp}

Uniform frame times need not map linearly to physiological phase. Heart rate affects systole and diastole differently, with substantially greater diastolic variability \citep{chung2004duration}. Thus, the same $t$ can represent different physiological states across subjects, potentially distorting learned dynamics and disease-related effects.
To address this, we introduce a learnable monotone phase reparameterization:
\begin{equation}
  \theta(t) = 2\pi\,\frac{\int_0^t \omega\bigl(\sin(2\pi s),\cos(2\pi s),\mathrm{HR}\bigr)\,ds}{\int_0^1 \omega\bigl(\sin(2\pi s),\cos(2\pi s),\mathrm{HR}\bigr)\,ds},
  \label{eq:warp}
\end{equation}
where $\mathrm{HR}$ denotes heart rate and $\omega(\cdot) > 0$ is a compact two-layer MLP (hidden size 16, tanh activations, softplus output). Positivity makes $\theta(t)$ monotonic, and normalization guarantees $\theta(0)=0$ and $\theta(1)=2\pi$.

The latent dynamics in Eq.~\eqref{eq:latent_ode} are driven by the \emph{warped} phase $\theta(t)$ rather than directly by $t$, allowing heart rate-dependent temporal stretching and compression. We restrict the warping to depend only on heart rate to separate exogenous timing variation from intrinsic cardiac dynamics.

\subsection{Mesh Likelihood with Graph Decoding}

For an observed frame $t$, we model the (standardized) bi-ventricular mesh $\mathbf{x}(t) \in \mathbb{R}^{3V}$ with a Gaussian likelihood, conditioned on the latent state $\mathbf{z}(t)$:
\begin{equation}
p\bigl(\mathbf{x}(t) \mid \mathbf{z}(t)\bigr)
= \mathcal{N}\bigl(\mathbf{g}(\mathbf{z}(t)), \mathbf{I}\bigr),
\end{equation}
where $\mathbf{g}(\cdot)$ is a three-layer hierarchical GNN decoder based on SpiralNet++ \citep{gong2019spiralnet++}. Because mesh connectivity is fixed across subjects and time, smooth latent trajectories map to anatomically coherent motion. We use 64 convolutional filters throughout.

\subsection{Variational Inference and Training Objective}
\label{subsec:var}

We train the model within a conditional VAE framework \citep{kingma2013auto,sohn2015learning}. Given an observed mesh sequence $\{\mathbf{x}(t_j)\}_{j=1}^{T}$ of $T$ frames, together with covariates $\mathbf{u}$ and heart rate $\mathrm{HR}$, the encoder infers an approximate posterior over the initial latent state:
\begin{equation}
q\bigl(\mathbf{z}(0) \mid \{\mathbf{x}(t_j)\}_{j=1}^{T}, \mathbf{u}, \mathrm{HR}\bigr),
\end{equation}
which determines the full trajectory through the latent ODE.

The encoder mirrors the decoder's hierarchical GNN at each frame and aggregates frame-wise embeddings with a two-layer transformer \citep{vaswani2017attention}. Attentive pooling of the transformer output frame tokens gives $\mathbf{z}(0)$, with covariates $\mathbf{u}$ added as auxiliary input tokens. Warped phase $\theta(t_j)$ defines each frame's (fixed) positional embedding, encoding periodic structure and heart rate effects (Section~\ref{subsec:phase_warp}): analogous to the standard transformer positional encoding, we concatenate $\sin(k\theta(t_j))$ and $\cos(k\theta(t_j))$ across integer harmonics $k=1,\dots,d/2$ to form a $d$-dimensional embedding added to each frame token.

Across a training set of $N$ subjects, the model is trained by maximizing the evidence lower bound (ELBO):
\begin{multline}
\mathcal{L}
= \sum_{i=1}^{N} \Biggl[ \sum_{j=1}^{T} \mathbb{E}_{q_i} \bigl[ \log p\left(\mathbf{x}_i(t_j) \mid \mathbf{z}_i(t_j)\right) \bigr] \\
- \beta\,\mathrm{KL}\Bigl(q_i(\mathbf{z}_i(0) \mid \cdot\,)\,\|\, p(\mathbf{z}_i(0) \mid \mathbf{u}_i)\Bigr) \Biggr],
\label{eq:loss_function}
\end{multline}
where $q_i$ is the approximate posterior for subject $i$, and $\beta$ balances reconstruction fidelity and prior regularization.

\paragraph{Relation to identifiable VAEs.}
Different latent representations can explain the same data equally. Identifiability limits this ambiguity to a restricted class of transformations, making it desirable for model interpretability. Identifiable VAEs (iVAEs) achieve such guarantees under specific assumptions by conditioning latent priors on observed auxiliary variables \citep{khemakhem2020variational}. Here, covariates $\mathbf{u}$ modulate the prior mean and variance, inducing structured, non-stationary variability. Although we do not establish formal identifiability, we observe empirically that conditioning the prior improves reproducibility of subject--subject latent similarity (Section~\ref{subsec:robustness}), rather than uniqueness or physiological interpretability of individual coordinates.

\subsection{Survival Risk Scoring from Residual Embeddings}
\label{subsec:risk_score}

For downstream survival modeling, we summarize each subject by the posterior mean $\bar{\mathbf{z}}(0)=\mathbb{E}_{q(\mathbf{z}(0)\mid\cdot)}[\mathbf{z}(0)]$. We then define residual embeddings by standardizing this mean relative to the covariate-conditioned prior:
\begin{equation}
\hat{\mathbf{z}} = \frac{ \bar{\mathbf{z}}(0) - \boldsymbol{\mu}(\mathbf{u}) }{ \boldsymbol{\sigma}(\mathbf{u}) }.
\end{equation}
The posterior mean gives a stable deterministic representation, and normalization removes expected covariate effects to yield dimensionless deviations.

We then extract a disease-specific scalar risk score using a penalized Cox proportional hazards model:
\begin{equation}
    h(\tau \mid \hat{\mathbf{z}}) =
    h_0(\tau)\exp\left(
    \boldsymbol{\alpha}^\top M(\hat{\mathbf{z}})
    \right),
\end{equation}
where $h(\tau \mid \hat{\mathbf{z}})$ is the hazard at follow-up time $\tau$, $h_0(\tau)$ is the baseline hazard, and $M(\cdot)$ is a spline basis expansion of the residual embedding coefficients. 
Each coordinate $\hat{z}_d$ is transformed with natural cubic splines with three degrees of freedom to allow nonlinear risk effects. We fit the model with penalized Cox regression using penalty parameter 0.1, selected by 5-fold cross-validation on the development cohort over a logarithmic grid ($10^{-3}$ to $10^{1}$) (Appendix~\ref{sec:appendix_e}).
The resulting linear predictor
\begin{equation}
\rlat =
\boldsymbol{\alpha}^\top M(\hat{\mathbf{z}})
\label{eq:risk_score}
\end{equation}
defines a scalar \emph{latent risk score}. Higher \rlat{} indicates higher predicted hazard and greater event risk from abnormal spatiotemporal motion.

\section{Experiments}\label{sec:experiments}

We experimentally evaluate four questions: (1) whether the latent residual representation improves incident HF prediction beyond clinical and imaging risk factors; (2) how the architecture compares with alternative spatial and temporal formulations and a state-of-the-art approach; (3) whether the model generates physiologically realistic, covariate-dependent cardiac variation; and (4) whether the representations are interpretable and stable. We first present the study design, then the primary endpoint, architectural analysis, interpretability, and robustness.

\subsection{Experimental Setup}
\label{subsec:exp_setup}

\paragraph{Cohort and split.}
The cohort included UK Biobank participants who underwent CMR \citep{petersen2016uk} at instance 2. From an initial sample of 84{,}086 participants, we excluded subjects who had withdrawn consent, lacked required long-axis (LAX) or short-axis (SAX) acquisitions, had corrupted raw files, failed case-level view selection because of incomplete SAX coverage or irreconcilable metadata, or failed mesh fitting because essential guide-point or slice-information files were absent or invalid. We also removed subjects with measured LV or RV ejection fraction below 10\% after manual inspection, as these cases likely reflect failed processing.

To study incident risk, we excluded participants with baseline CVD or previous HF, mirroring the approach used in the pooled cohort equations to prevent heart failure (PCP-HF) \citep{khan201910}. Prevalent CVD used the composite outcome of \citet{elliott2020predictive}, augmented with peripheral artery disease as in \citep{klarin2019genome}; Appendix~\ref{sec:appendix_a} lists all UK Biobank fields and diagnostic codes. The final cohort comprised 72{,}386 participants, including 367 participants with incident HF events.

We randomly split the cohort 1:1 into a development and an evaluation cohort, stratified by incident HF status. All fitting (VAE training, spline Cox modeling for \rlat{}, missing-covariate imputation, and conventional CMR comparison models) used only the development cohort. All performances in this paper are reported on the held-out evaluation cohort.

\paragraph{Clinical endpoints.}
Incident heart failure is the primary endpoint because it is directly linked to abnormalities in ventricular structure and motion. Although clinically heterogeneous, HF is strongly coupled to remodeling, systolic dysfunction, diastolic dysfunction, or combinations thereof \citep{meunier2021202,heidenreich20222022}. It therefore provides a clear test of whether latent cardiac anatomy and dynamics capture disease-relevant variation.

We also evaluate secondary endpoints to assess broader cardiovascular relevance: cardiomyopathy\footnote{Although cardiomyopathy (similarly to HF) is an umbrella term covering various pathologies with very different disease courses, we treat it here as a single endpoint because of the limited number of occurrences.} and atrial fibrillation as remodeling-related outcomes; and cardiovascular and all-cause mortality as broader prognostic outcomes. Appendix~\ref{sec:appendix_a} lists all endpoint definitions and counts.

\paragraph{Mesh preprocessing, conventional CMR indices, and covariates.}
Bi-ventricular 3D+t meshes were extracted from cine CMR volumes using the open-source pipeline of \citet{dillon2025open}: automated segmentation \citep{isensee2021nnu}, contour and landmark extraction, breathhold correction \citep{sinclair2017fully}, and iterative diffeomorphic mesh fitting \citep{mauger2018iterative}. To improve biobank-scale robustness, we added three steps. First, quality control filters slice- and frame-level segmentations for anatomical plausibility and temporal consistency, including missing structures, abrupt area changes, excessive centroid displacement, fragmentation, low inter-frame overlap, and loss of cavity--myocardium adjacency. Second, cyclic Savitzky--Golay filtering with limited gap interpolation and outlier rejection smooths fitted mesh sequences, reducing frame-to-frame noise while preserving cyclic motion. Third, near-duplicate boundary vertices are collapsed to combine the LV endocardial, RV endocardial, and epicardial surfaces into a single watertight mesh.
The resulting representation has 5{,}806 vertices and consistent topology across subjects. After partial Procrustes alignment, all meshes are standardized using the population template mean and standard deviation.

From these meshes, we extract seven conventional CMR markers: left and right ventricular ejection fraction (LVEF and RVEF), LV global and RV free-wall longitudinal strain (computed by arc-length stretching of a predefined endocardial arc), left ventricular mass index, left ventricular wall thickness (average mid-cavity wall thickness), and relative wall thickness (LV wall thickness divided by average LV internal diameter). Some commonly used functional measures, such as the E/e' ratio, are not included because they require Doppler echocardiographic measurements and cannot be determined from conventional cine CMR alone.

For the covariate-conditioned prior, auxiliary variables $\mathbf{u}$ are age, sex, and BSA, computed from height and weight using the Du Bois formula \citep{du1916clinical}. Heart rate for phase warping is read from the DICOM header. Table~\ref{tab:tbl1} summarizes the clinical covariates used by the refitted PCP-HF model \citep{khan201910}. Because glucose and lipid measurements were not collected during imaging, we use the most recent preceding UK Biobank assessment. Casual glucose proxies fasting glucose, and missing clinical values are imputed with MICE \citep{van2011mice}. Unlike the original PCP-HF model, we do not include race.

\paragraph{Metrics.}
We assess the latent ODE model from three complementary perspectives: reconstruction fidelity, generative realism, and downstream survival prediction. This separation matters because accurate reconstruction, realistic population-level sampling, and prognostic utility need not coincide.

Reconstruction fidelity is measured by mean vertex-wise Euclidean distance, mean surface-normal angle error, and mean vertex acceleration error \citep{kanazawa2019learning}, capturing positional accuracy, local geometric agreement, and temporal consistency. Generative realism is evaluated on 1{,}000 samples (conditioned on real evaluation cohort covariates) using the mean Wasserstein distance between distributions of the seven CMR indices (each normalized by empirical standard deviation), minimum matching distance (MMD) \citep{achlioptas2018learning} to the nearest real mesh (based on five subsampled frames for efficiency), and temporal smoothness from mean vertex acceleration $\mathbf{x}(t_{j+1}) - 2\mathbf{x}(t_j) + \mathbf{x}(t_{j-1})$. For clinical utility, we report sex-stratified survival metrics: primarily HF Harrell's C-index \citep{harrell1982evaluating}, and where relevant time-dependent AUROC and calibration by expected/observed (E/O) ratio. Metrics are computed within sex strata and aggregated by pair-count pooling for discrimination and pooled expected and observed counts for calibration.

\paragraph{Implementation.}
The GNN+ODE model is implemented in \software{PyTorch} \citep{paszke2019pytorch}, with the latent ODE solved by an adaptive-step Dormand--Prince integrator \citep{dormand1980family}. Unless stated otherwise, $D=64$ and $\beta=5.0$ in Eq.~\eqref{eq:loss_function}. We train with Adam \citep{kingma2014adam} using batch size 8, learning rate $10^{-3}$, weight decay $10^{-6}$, and validation-based early stopping. Survival analysis uses \software{lifelines} \citep{DavidsonPilon2019}. Code is available at \url{https://github.com/brdav/cardiac_latent_ode}.

\begin{table*}[width=.95\textwidth,cols=6,pos=h]
\caption{Baseline cohort characteristics and PCP-HF covariates, reported for the full study population and stratified by incident heart failure (HF) status. Values are shown as mean (SD) or count (\%), and the number of missing observations is provided for each variable.}\label{tab:tbl1}
\centering
\footnotesize
\setlength{\tabcolsep}{4pt}
\renewcommand{\arraystretch}{1.05}
\begin{tabular*}{\tblwidth}{@{}LLRRR@{}}
\toprule
 & Missing, n & Overall & No incident HF & Incident HF \\ 
\midrule
Total, n & - & 72386 & 72019 & 367 \\
\midrule
Mean age, years (SD) & 0 & 65.7 (7.9) & 65.6 (7.9) & 69.2 (6.8) \\
Female, n (\%) & 0 & 39112 (54.0) & 38979 (54.1) & 133 (36.2) \\
Casual glucose, mg/dL (SD) & 9469 & 89.5 (16.1) & 89.5 (16.1) & 92.5 (18.9) \\
Diabetes treatment, n (\%) & 872 & 1111 (1.5) & 1097 (1.5) & 14 (3.8) \\
Current smoking, n (\%) & 21 & 2363 (3.3) & 2350 (3.3) & 13 (3.5) \\
Mean systolic blood pressure, mmHg (SD) & 10 & 140.8 (19.4) & 140.7 (19.4) & 144.4 (18.1) \\
Hypertension treatment, n (\%) & 23 & 17791 (24.6) & 17639 (24.5) & 152 (41.4)\\
Mean total cholesterol, mg/dL (SD) & 3917 & 223.0 (41.1) & 223.1 (41.1) & 215.9 (42.5) \\
Mean HDL cholesterol, mg/dL (SD) & 9411 & 57.9 (14.6) & 57.9 (14.6) & 54.0 (14.6) \\
Mean body mass index, kg/m$^2$ (SD) & 0 & 26.1 (4.4) & 26.1 (4.4) & 27.5 (4.9) \\
Mean QRS duration, ms (SD) & 7297 & 87.7 (14.1) & 87.6 (14.0) & 95.9 (20.9) \\
\bottomrule
\end{tabular*}
\end{table*}

\subsection{Primary Endpoint: Incident Heart Failure}
\label{subsec:primary_endpoint}

\begin{table*}[width=.95\textwidth,pos=h]
\caption{Prediction of incident heart failure in evaluation cohort when augmenting the PCP-HF variables with the conventional CMR-index score \rconv{} and/or the latent score \rlat{} (defined in Sec.~\ref{subsec:risk_score} and \ref{subsec:primary_endpoint}). Higher C-index and AUROC and an E/O ratio closer to 1 indicate better performance. Metrics are stratified by sex. Values are reported as point estimates $\pm$ bootstrap-estimated standard deviations from 1{,}000 resamples.}\label{tab:comparison_image_markers}
\centering
\footnotesize
\setlength{\tabcolsep}{4pt}
\renewcommand{\arraystretch}{1.05}
\begin{tabular*}{\tblwidth}{@{} Lcccccccc@{} }
\toprule
\multirow{2}{*}{Predictors} && \multirow{2}{*}{C-index} && \multicolumn{2}{c}{3-Year Horizon} && \multicolumn{2}{c}{5-Year Horizon}\\
\cmidrule{5-6} \cmidrule{8-9}
&&&& AUROC & E/O Ratio && AUROC & E/O Ratio \\
\midrule
PCP-HF && $0.704 \pm 0.021$ && $0.671 \pm 0.028$ & $\mathbf{0.977 \pm 0.112}$ && $0.661 \pm 0.023$ & $1.228 \pm 0.106$ \\
PCP-HF + \rconv{} && $0.764 \pm 0.020$ && $0.747 \pm 0.029$ & $0.948 \pm 0.109$ && $0.728 \pm 0.023$ & $1.210 \pm 0.105$ \\
PCP-HF + \rlat{} && $\mathbf{0.785 \pm 0.019}$ && $\mathbf{0.778 \pm 0.028}$ & $0.927 \pm 0.106$ && $\mathbf{0.751 \pm 0.023}$ & $1.163 \pm 0.101$ \\
PCP-HF + \rconv{} + \rlat{} && $0.784 \pm 0.020$ && $0.775 \pm 0.028$ & $0.915 \pm 0.105$ && $0.750 \pm 0.023$ & $\mathbf{1.157 \pm 0.100}$ \\
\bottomrule
\end{tabular*}
\end{table*}

\begin{figure}
  \centering
  \includegraphics[width=\linewidth]{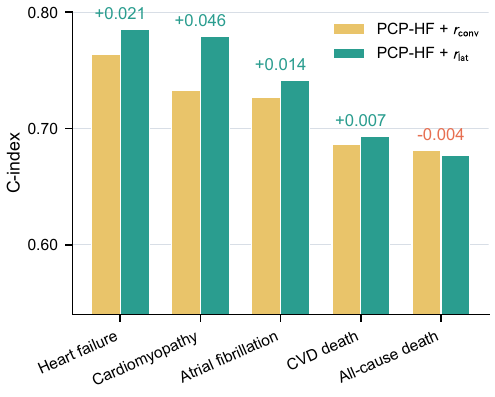}
    \caption{Sex-stratified C-indices in evaluation cohort for secondary cardiovascular endpoints, using the conventional CMR-index score \rconv{} or the latent score \rlat{}, in addition to PCP-HF variables.}\label{fig:secondary_comparison}
\end{figure}

Incident HF prediction is the primary downstream evaluation. As a clinical comparator, we refit PCP-HF on the development cohort to account for variable differences (Section~\ref{subsec:exp_setup}). Rather than fully separate sex-specific models, we use shared covariate coefficients with sex-specific baseline hazards. This refitting improves the sex-stratified C-index from 0.685 with published coefficients to 0.704 on our data.

We augment PCP-HF with two imaging risk scores: the latent score \rlat{} (Eq.~\eqref{eq:risk_score}) and a conventional CMR-index score, \rconv{}. The latter fits the same spline Cox model (Section~\ref{subsec:risk_score}) to the seven CMR indices from Section~\ref{subsec:exp_setup} and auxiliary variables $\mathbf{u}$.

Table~\ref{tab:comparison_image_markers} shows that both imaging scores improve prediction over PCP-HF alone, with a larger gain from \rlat{}. Adding \rlat{} increases the C-index to 0.785, compared with 0.764 for \rconv{}. This C-index difference is statistically significant in a paired-bootstrap comparison (95\% CI: $0.0056$ to $0.0371$, $p=0.008$). The same pattern holds for 3- and 5-year AUROC. In the PCP-HF~+~\rlat{} model, the hazard ratio per standard deviation increase in \rlat{} is 1.51 (95\% CI: 1.43--1.59, $p=1.15\times10^{-51}$).
Calibration by E/O ratio is mixed: PCP-HF alone is closest to 1 at 3 years (0.977), whereas imaging scores move the 5-year baseline E/O of 1.228 closer to 1, especially with \rlat{} (1.163). Thus, \rlat{} gives the strongest discrimination gains, but not uniformly better calibration.
Combining \rconv{} with \rlat{} did not increase the C-index over \rlat{} alone (cf.\ rows 3 and 4 of Table~\ref{tab:comparison_image_markers}).

Figure~\ref{fig:secondary_comparison} extends the comparison to secondary endpoints by evaluating \rconv{} and \rlat{} alongside PCP-HF predictors. The latent score performs best for HF, cardiomyopathy, and atrial fibrillation, consistent with the expectation that cine anatomy and motion capture remodeling-related processes most directly. For mortality outcomes, performance is broadly comparable between the two imaging scores, likely reflecting the broader influence of vascular and systemic risk factors beyond ventricular motion.

\subsection{Architectural Comparison}

\begin{table*}[width=.95\textwidth,cols=8,pos=h]
\caption{Architectural comparison of the proposed model and benchmark models in evaluation cohort. Lower is better for reconstruction and generative metrics; higher is better for risk prediction. Values are reported as point estimates $\pm$ bootstrap-estimated standard deviations from 1{,}000 resamples. Model definitions are in Appendix~\ref{sec:appendix_b}. ODE: ordinary differential equation, GNN: graph neural network, (uncond.): model does not implement conditional generation.}\label{tab:baseline_comparison}
\centering
\footnotesize
\setlength{\tabcolsep}{4pt}
\renewcommand{\arraystretch}{1.05}
\begin{tabular*}{\tblwidth}{@{\extracolsep{\fill}}lccccccc@{}}
\toprule
\multirow{2}{*}{Model} & \multicolumn{3}{c}{Reconstruction Error} & \multicolumn{3}{c}{Generative Quality} & Risk Prediction \\
\cmidrule(lr){2-4} \cmidrule(lr){5-7} \cmidrule(l){8-8}
 & Eucl. (mm) & Ang. (deg) & Acc. (mm) & Wasserst. & MMD (mm) & Smooth (mm) & C-index \\
\midrule
Linear+ODE & $1.548 \pm 0.002$ & $13.316 \pm 0.009$ & $0.502 \pm 0.001$ & $0.176 \pm 0.025$ & $\mathbf{2.915 \pm 0.013}$ & $0.134 \pm 0.001$ & $0.768 \pm 0.021$ \\
GNN+Fourier & $2.050 \pm 0.002$ & $14.443 \pm 0.007$ & $0.506 \pm 0.001$ & $\mathbf{0.120 \pm 0.028}$ & $5.320 \pm 0.018$ & $\mathbf{0.124 \pm 0.001}$ & $0.761 \pm 0.020$ \\
GNN+Transformer & $1.693 \pm 0.002$ & $15.766 \pm 0.007$ & $1.179 \pm 0.001$ & $0.198 \pm 0.025$ & $3.445 \pm 0.033$ & $1.033 \pm 0.005$ & $0.778 \pm 0.020$ \\
\midrule
4D CardioSynth & $2.003 \pm 0.002$ & $14.902 \pm 0.010$ & $0.534 \pm 0.001$ & \emph{(uncond.)} & \emph{(uncond.)} & $0.153 \pm 0.001$ & $0.760 \pm 0.020$ \\
MeshHeart & $1.614 \pm 0.001$ & $15.340 \pm 0.007$ & $0.818 \pm 0.002$ & $0.199 \pm 0.026$ & $3.082 \pm 0.023$ & $0.274 \pm 0.003$ & $0.779 \pm 0.019$ \\
\midrule
GNN+ODE & $\mathbf{1.366 \pm 0.001}$ & $\mathbf{12.561 \pm 0.008}$ & $\mathbf{0.498 \pm 0.001}$ & $0.141 \pm 0.025$ & $3.007 \pm 0.011$ & $0.152 \pm 0.001$ & $\mathbf{0.785 \pm 0.019}$ \\
\bottomrule
\end{tabular*}
\end{table*}

Having established the primary downstream benefit, we compare the proposed model with alternative architectures. Table~\ref{tab:baseline_comparison} benchmarks combinations of linear or graph-based mesh decoders with ODE-, Fourier-, and transformer-based temporal models. We report the metrics from Section~\ref{subsec:exp_setup} to compare subject-level reconstruction, population-level generative realism, and downstream HF discrimination.
We also include 4D CardioSynth \citep{dou20254d} and MeshHeart \citep{qiao2025personalized}, two published benchmark architectures that combine graph-based spatial representations with disentangled recurrent dynamics and transformer-based temporal modeling, respectively. 

GNN+ODE achieves the best reconstruction accuracy across all three reconstruction metrics and the highest HF C-index. Linear spatial encoders/decoders reconstruct worse and predict less well, indicating that mesh connectivity provides useful anatomical structure. Transformer- and Fourier-based temporal models also underperform the ODE on downstream HF prediction, as does 4D CardioSynth. For 4D CardioSynth, this may reflect its sole reliance on the end-diastolic (ED) frame for encoding: although the model generates full-cycle motion, its embedding does not incorporate observed motion across the cardiac cycle. MeshHeart produces smoother trajectories than the plain transformer variant, likely because of Laplacian smoothness regularization, but this does not substantially improve HF prediction or reconstruction.
Compared with the best-performing benchmark MeshHeart, the C-index difference is not statistically significant in bootstrapped samples (95\% CI: $-0.0049$ to $0.0227$, $p=0.210$), likely reflecting the small number of incident HF cases. However, GNN+ODE is significantly better in Euclidean reconstruction error ($p=5.53\times10^{-5}$) and Wasserstein distance ($p=4.00\times10^{-3}$). Although some variants score slightly better on MMD, these gains do not coincide with better reconstruction or risk prediction. We therefore interpret GNN+ODE as the best trade-off between anatomical fidelity, temporal consistency, and prognostic utility.

\subsection{Physiological Validity of the Generative Model}

Table-level distributional metrics give only a coarse view of generative quality. We therefore test whether the conditional prior reproduces known age- and sex-dependence in cardiac phenotype. We sample covariates from the study population, generate full cardiac-cycle meshes, compute clinical indices, and compare age- and sex-stratified trends with empirical observations.

\begin{figure}
  \centering
  \includegraphics[width=\linewidth]{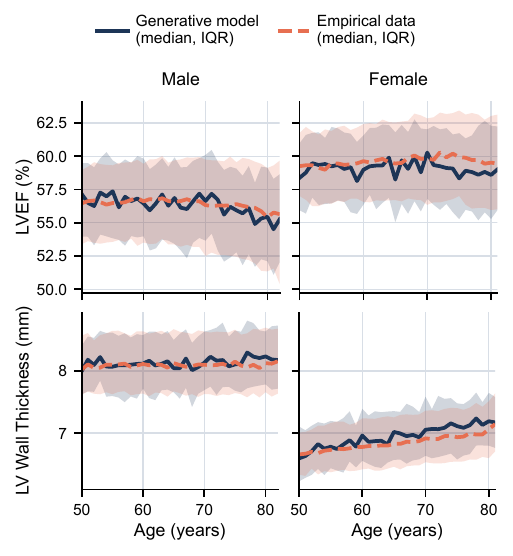}
    \caption{Subgroup analysis: age- and sex-stratified comparison of LVEF (top) and LV wall thickness (bottom) between generative-model samples (solid blue line, shaded interquartile range (IQR)) and the empirical cohort (dashed orange line, shaded IQR). Males are shown on the left and females on the right. Agreement in central tendency and dispersion supports the realism of the covariate-conditioned generative model.}\label{fig:covariate_traversal_main}
\end{figure}

Figure~\ref{fig:covariate_traversal_main} shows close agreement between generated and empirical medians and interquartile ranges for LVEF and LV wall thickness in men and women. The model reproduces age-related attenuation of LVEF in males and increased wall thickness in females. Together with Table~\ref{tab:baseline_comparison}, this suggests that the covariate-conditioned prior captures physiologically realistic variation rather than merely plausible meshes. Appendix~\ref{sec:appendix_d} shows additional indices.

\subsection{Interpreting the Learned Risk Axis}
\label{subsec:interpretability}

To interpret the learned risk score, we compare two latent states for each subject: the covariate-conditioned prior mean, representing expected anatomy given age, sex, and BSA, and the posterior mean, obtained from the observed mesh sequence. Decoding both states yields two full-cycle mesh sequences; their difference highlights subject-specific remodeling beyond the covariates.

For visualization, we select representative high-risk cases. We take the top 1\% of evaluation subjects by predicted HF risk, stratify by sex, and split each sex into preserved-EF (LVEF $\geq 50\%$) and reduced-EF (LVEF $< 50\%$) groups. Within each subgroup, we choose the subject closest to the subgroup centroid to highlight typical high-risk phenotypes rather than outliers.

\begin{figure}
  \centering
  \includegraphics[width=\linewidth]{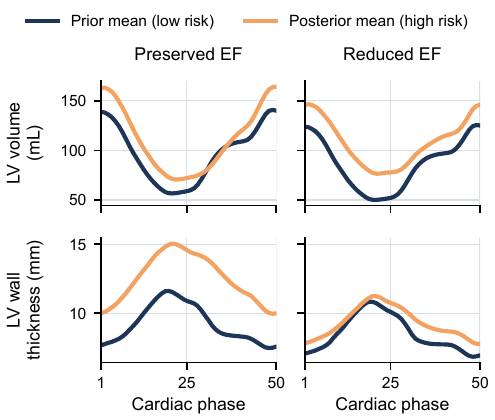}
    \caption{Comparison of the covariate-conditioned prior-mean and posterior-mean reconstructions for two representative high-risk female subjects. Left: preserved ejection fraction. Right: reduced ejection fraction. Top row: LV volume curves. Bottom row: LV wall thickness. Blue denotes the prior mean and orange the posterior mean inferred from the observed mesh sequence. In the preserved-EF case, the posterior is dominated by wall thickening with impaired diastolic relaxation, whereas in the reduced-EF case it is dominated by ventricular dilation.}\label{fig:volume_curves}
\end{figure}

Figure~\ref{fig:volume_curves} compares prior- and posterior-mean reconstructions for two representative female subjects. In the preserved-EF case, the posterior mainly thickens the LV wall throughout the cycle and accentuates diastolic dysfunction, consistent with concentric remodeling. In the reduced-EF case, the posterior shifts the LV volume curve upward, indicating dilation. Thus, posterior departures from the covariate-conditioned prior capture distinct routes to HF: elevated risk does not imply a single remodeling pattern. Appendix~\ref{sec:appendix_d} provides equivalent plots for males.

\begin{figure*}
  \centering
  \includegraphics[width=\linewidth]{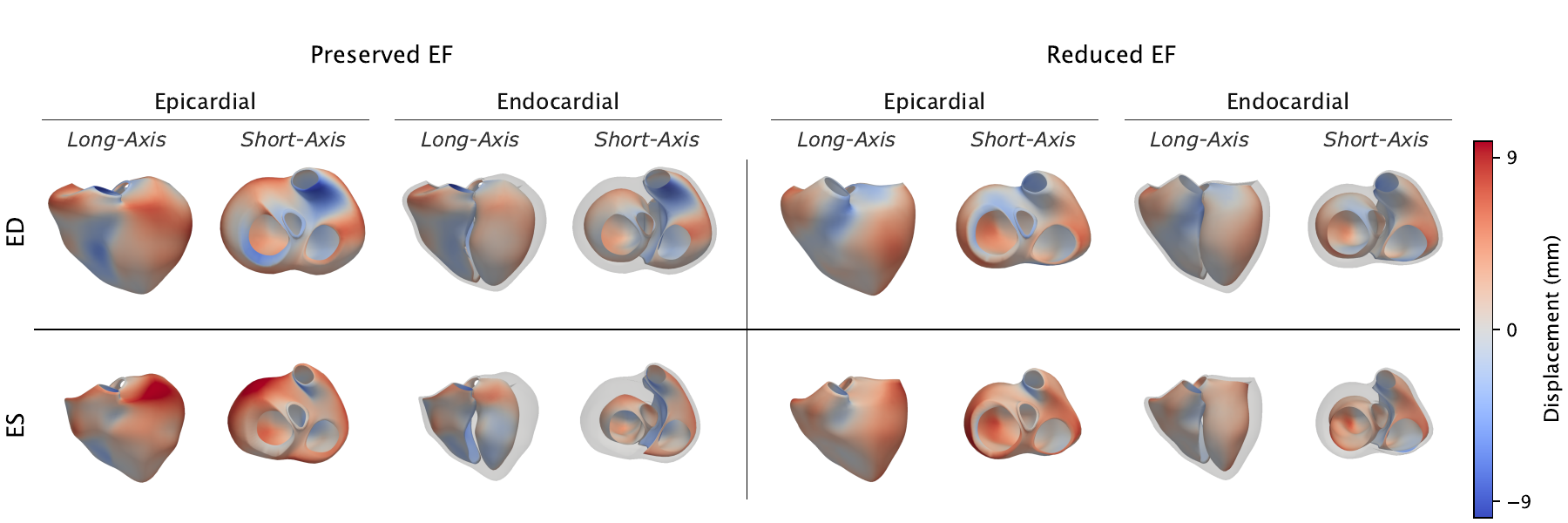}
    \caption{Surface displacement of the mesh decoded from the posterior mean relative to the subject-specific prior mean for the two female subjects in Figure~\ref{fig:volume_curves}. Rows show end diastole (ED) and end systole (ES); within each subject, columns show epicardial and endocardial surfaces rendered in two orthogonal orientations: a long-axis and short-axis view. Red indicates outward displacement and blue inward displacement along the surface normal. The preserved-EF case shows predominantly epicardial expansion consistent with wall thickening, whereas the reduced-EF case shows more global outward displacement consistent with ventricular dilation.}\label{fig:mesh_vis}
\end{figure*}

Figure~\ref{fig:mesh_vis} supports this interpretation. In the preserved-EF case, displacement is concentrated on the epicardial surface, especially at end-systole, matching the wall-thickening pattern in Figure~\ref{fig:volume_curves}. In the reduced-EF case, both endocardial and epicardial surfaces move outward diffusely at end-diastole and end-systole, consistent with global dilation. Appendix~\ref{sec:appendix_d} shows corresponding male visualizations.

\subsection{Component Analysis}

\begin{figure}
    \centering
    \includegraphics[width=\linewidth]{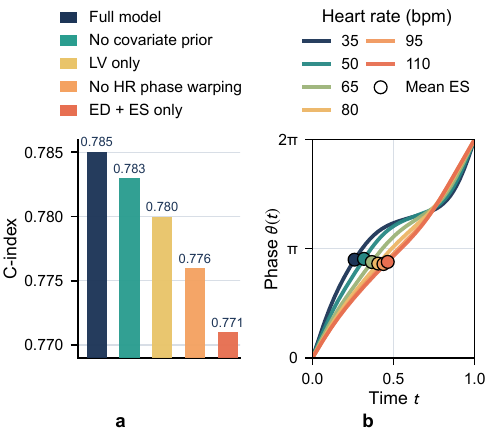}
    \caption{Component analysis of the proposed model. Panel \textbf{a} ablates important model components for HF prediction. Panel \textbf{b} visualizes the learned phase trajectories and shows improved alignment of physiological phase across subjects after warping. ED: end diastole, ES: end systole, HR: heart rate, LV: left ventricle.}
    \label{fig:mechanism}
\end{figure}

We next test whether the full model's gains come from the intended architectural components. Figure~\ref{fig:mechanism} combines an ablation study with direct inspection of phase warping.

Panel~\textbf{a} shows that the temporal ablations produce the largest observed performance decreases. Removing covariate conditioning has little effect, and excluding the right ventricle modestly reduces HF prediction performance, suggesting an incremental contribution of RV information. Removing heart-rate-aware phase warping lowers the C-index by 0.009, while restricting the model to end-diastolic and end-systolic frames produces the largest drop ($\Delta$C-index $=0.0140$). A paired bootstrap result for the latter was at the margin of statistical significance at the 0.05 level (95\% CI: $-0.0001$ to $0.0273$, $p=0.054$).

Panel~\textbf{b} confirms that phase warping behaves as intended. After warping, average end-systole aligns more tightly across subjects despite heart-rate variability, suggesting that the learned warping captures a valid physiological reparameterization of time.

\subsection{Robustness and Sensitivity}
\label{subsec:robustness}

\begin{figure}
    \centering
    \includegraphics[width=\linewidth]{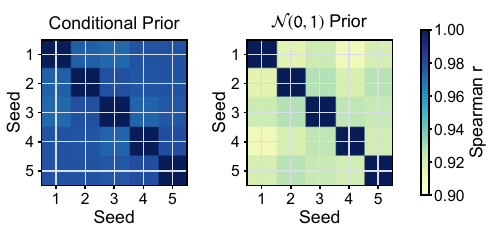}
    \caption{To evaluate latent robustness we compare across-seed representational similarity for the full model (left) and the ablation using a standard normal prior (right).}
    \label{fig:latent_robustness}
\end{figure}

\begin{figure}
    \centering
    \includegraphics[width=\linewidth]{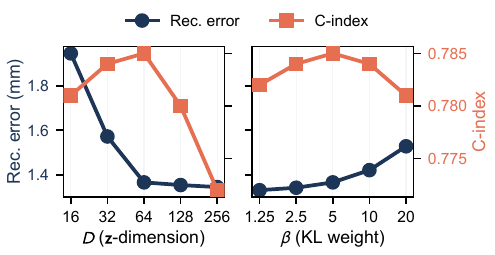}
    \caption{Sensitivity analysis with respect to the latent dimension $D$ and KL divergence weight $\beta$.}
    \label{fig:sensitivity}
\end{figure}

Figure~\ref{fig:latent_robustness} evaluates latent stability across random initializations by comparing the full model with an ablation lacking a covariate-conditioned prior. For each variant, we train five seeds and perform representational similarity analysis (RSA) \citep{kriegeskorte2008representational} using Spearman correlations between subject--subject latent similarity matrices across runs. The covariate-conditioned prior gives higher cross-seed consistency than the unconditioned ablation (mean pairwise Spearman $r=0.977$ versus $0.922$). Thus, even with modest discrimination gains, conditioning substantially improves latent reproducibility.

Figure~\ref{fig:sensitivity} shows sensitivity to latent dimension $D$ and KL weight $\beta$. As expected, reconstruction error decreases as $D$ increases and $\beta$ decreases. However, prediction peaks near the selected configuration, $D=64$ and $\beta=5.0$, indicating that moderate regularization helps prognosis even when it does not minimize reconstruction loss.

Because only the posterior mean $\bar{\mathbf{z}}(0)$ is used in the Cox stage (Section~\ref{subsec:risk_score}), we additionally assessed the effect of propagating the full posterior at inference (Appendix~\ref{sec:appendix_g}). Averaging \rlat{} over $K=50$ posterior samples per subject changes discrimination only slightly (C-index 0.785 for the point estimate versus 0.779 for the sample-averaged score). However, the per-subject posterior standard deviation of \rlat{} varies across the cohort and is positively correlated with the point-estimate risk score (Spearman $r=0.206$), indicating that subjects assigned higher risk tend to have greater posterior uncertainty. As a point estimate, our risk score does not communicate this subject-specific uncertainty, as further discussed in Appendix~\ref{sec:appendix_g}.

\section{Discussion and Limitations}\label{sec:discussion}

Mesh-based continuous full-cycle modeling of ventricular motion provides informative cardiac phenotypes beyond conventional CMR summaries. For incident HF prediction, the latent score reached a C-index of 0.785 in a held-out UK Biobank evaluation cohort. Across architectural comparisons, the GNN+ODE model also gave the strongest trade-off between reconstruction quality, generative realism, and downstream HF discrimination. Importantly, the mesh-based representation remains anatomically interpretable, enabling investigation of regional and temporal patterns of cardiac remodeling and motion abnormalities that may underlie future cardiovascular disease.

The model learns clinically meaningful structure rather than arbitrary mesh compression: the covariate-conditioned prior reproduced known age- and sex-related ventricular trends \citep{ji2022sex}, while residual embeddings yielded plausible subject-specific deviations. The learned risk axis did not map to a single remodeling pattern: selected high-risk subjects showed wall thickening with impaired relaxation \citep{aziz2013diastolic} in preserved-EF cases and dilation in reduced-EF cases. Because HF spans multiple remodeling trajectories \citep{meunier2021202,heidenreich20222022}, such shared latent representations may suit population-level risk modeling better than a single phenotypic axis.

Ablation and robustness analyses highlight the value of the model design choices. Covariate conditioning, heart-rate-aware phase warping, and full-cycle modeling each improved either prediction or reproducibility. In particular, the conditioned prior increased latent stability, consistent with identifiability benefits from auxiliary-variable conditioning in variational models \citep{khemakhem2020variational}. This matters if the embedding is to serve not only as a predictor but also as a reproducible representation of cardiac remodeling.

Several limitations remain. First, the framework depends on the upstream mesh extraction pipeline \citep{dillon2025open}. Although we add quality control and temporal smoothing, template-based meshes may inherit segmentation and registration biases. Diffeomorphic fitting is also performed independently at each time point before smoothing, so temporal coherence is not enforced during mesh construction. Future work could use joint spatiotemporal fitting \citep{zhang2010cardiac4d,liu2017spatiotemporal}, uncertainty-aware segmentation and mesh extraction \citep{guo2022uncertaintyshape,ng2023uncertainty}, or end-to-end learning from images \citep{kong2021wholeheart,mercadier2025printmesh}.

Second, as discussed in Section~\ref{subsec:latent_dynamics}, the latent ODE does not explicitly enforce periodicity. Although a simple cycle-consistency penalty did not improve modeling fidelity or downstream performance, richer periodic structure---such as explicit periodic latent parameterizations---may improve dynamical faithfulness and stability. Similarly, the first-order ODE is a flexible phenomenological model rather than a mechanistic one. Higher-order or biomechanically informed systems may better capture myocardial motion by modeling acceleration, elastic recoil, or other physical temporal effects \citep{wang2009modelling,yildiz2019ode2vae,greydanus2019hamiltonian}.

Third, the survival analysis should be viewed as prognostic evaluation within UK Biobank rather than as a deployment-ready HF prediction model. Although the imaging cohort is large, only 367 incident HF events were observed, and UK Biobank participants are healthier and less socioeconomically representative than the general population \citep{fry2017comparison}. The learned score demonstrates that the latent representation contains prognostic signal in this setting, but it does not establish calibration or clinical utility in higher-risk patient populations. Validation in external cohorts and registries, such as disease-enriched CMR databases or representative population studies, would require robust mesh extraction across scanners and acquisition protocols, recalibration of the covariate-conditioned prior for the target population, and recalibration of the survival model for the target cohort's baseline risk and case mix.

Finally, the framework focuses on ventricular anatomy. This is appropriate for HF, in fact strong performance is retained even when excluding the RV and relying solely on spatiotemporal LV modeling (consistent with recent findings \citep{chadalavada2025mri}). However, the model omits structures that may carry complementary information, including the atria, valves, and great vessels. Extending toward a whole-heart representation could broaden utility for other cardiovascular phenotypes and further improve risk stratification \citep{muffoletto2024evaluation}.

\section{Conclusion}\label{sec:conclusion}

We presented a probabilistic framework for full-cycle cine CMR modeling that combines mesh-based graph representations, continuous latent ODE dynamics, heart-rate-aware phase warping, and a covariate-conditioned prior. In UK Biobank, the residual latent representation provided prognostic information for incident heart failure beyond refitted clinical risk equations and selected established CMR markers, while producing physiologically realistic samples and interpretable subject-specific remodeling patterns.

More broadly, these results suggest that spatiotemporal representations of the full cardiac motion cycle can reveal clinically relevant phenotypes missed by static summaries.
Future work should prioritize external validation in more representative and disease-enriched cohorts, richer dynamical structure through explicit periodic or higher-order models, and extension to more complete cardiac anatomy.


\appendix
\makeatletter
\@addtoreset{figure}{section}
\@addtoreset{table}{section}
\makeatother
\setcounter{figure}{0}
\setcounter{table}{0}
\renewcommand{\thefigure}{\thesection.\arabic{figure}}
\renewcommand{\thetable}{\thesection.\arabic{table}}

\section{Variable Definitions}
\label{sec:appendix_a}

This section lists the code lists and UK Biobank fields used for cohort construction and endpoint definition. Table~\ref{tab:prevalent} reports the ICD-10, ICD-9, OPCS-4, and additional fields used to identify prevalent cardiovascular disease (CVD) and peripheral artery disease (PAD) at baseline for exclusion. Table~\ref{tab:endpoint_def} summarizes the UK Biobank sources and coding definitions for heart failure and secondary endpoints.

\begin{table*}[width=.95\textwidth]
    \caption{UK Biobank fields used to define prevalent cardiovascular disease (CVD) and peripheral artery disease (PAD).}
    \label{tab:prevalent}
    \centering
    \footnotesize
    \setlength{\tabcolsep}{4pt}
    \renewcommand{\arraystretch}{1.05}
    \begin{tabular}{@{}>{\raggedright\arraybackslash}p{0.11\textwidth}>{\raggedright\arraybackslash}p{0.16\textwidth}>{\raggedright\arraybackslash}p{0.10\textwidth}>{\raggedright\arraybackslash}p{0.22\textwidth}>{\raggedright\arraybackslash}p{0.29\textwidth}@{}}
    \toprule
    \textbf{Definition} & \textbf{ICD-10} & \textbf{ICD-9} & \textbf{OPCS-4} & \textbf{Other UK Biobank fields} \\
    \midrule
    CVD \citep{elliott2020predictive} &
    G45, I20, I21, I22, I23, I24, I25, I63, I64 &
    410, 411, 412, 413, 414, 434, 436 &
    K40, K41, K42, K43, K44, K45, K46, K47.1, K49, K50, K75 &
    Field 20002: 1075, 1082, 1583; \newline Field 20004: 1070, 1071, 1105, 1109, 1514; \newline Field 6150: 1, 2, 3 \\
    \addlinespace[3pt]
    PAD \citep{klarin2019genome} &
    I70.0, I70.00, I70.01, I70.2, I70.20, I70.21, I70.8, I70.80, I70.9, I70.90, I73.8, I73.9 &
    4400, 4402, 4438, 4439 &
    X09.3, X09.4, X09.5, L21.6, L51.3, L51.6, L51.8, L52.1, L52.2, L54.1, L54.4, L54.8, L59.1, L59.2, L59.3, L59.4, L59.5, L59.6, L59.7, L59.8, L60.1, L60.2, L63.1, L63.5, L63.9, L66.7 &
    Field 20002: 1067, 1087, 1088; \newline Field 20004: 1102, 1108, 1440 \\
    \bottomrule
    \end{tabular}
\end{table*}

\begin{table*}[width=.95\textwidth]
    \caption{UK Biobank data sources used to define study endpoints, along with number of occurrences n.}
    \label{tab:endpoint_def}
    \centering
    \footnotesize
    \setlength{\tabcolsep}{4pt}
    \renewcommand{\arraystretch}{1.05}
    \begin{tabular}{@{}>{\raggedright\arraybackslash}p{0.22\textwidth}>{\raggedright\arraybackslash}p{0.15\textwidth}>{\raggedright\arraybackslash}p{0.16\textwidth}>{\raggedright\arraybackslash}p{0.20\textwidth}>{\raggedright\arraybackslash}p{0.15\textwidth}@{}}
    \toprule
    \textbf{Endpoint (n)} & \makecell[l]{\textbf{First occurrences} \\ \textbf{(Category 1712)}} & \makecell[l]{\textbf{Summary diagnoses} \\ \textbf{(Category 2002)}} & \makecell[l]{\textbf{Algorithmically-defined} \\ \textbf{outcomes (Category 42)}} & \makecell[l]{\textbf{Death register} \\ \textbf{(Category 100093)}} \\
    \midrule
    Heart failure (367) & I50 & I11.0, I13.0, I13.2 & -- & -- \\
    Cardiomyopathy (62) & I42 & -- & -- & -- \\
    Atrial fibrillation (995) & I48 & -- & -- & -- \\
    CVD death (212) & -- & -- & -- & ICD-10 chapter I \\
    All-cause death (1182) & -- & -- & -- & All causes \\
    \bottomrule
    \end{tabular}
\end{table*}

\section{Benchmark Model Definitions}
\label{sec:appendix_b}

Table~\ref{tab:baseline_comparison} compares the proposed GNN+ODE model with variants that isolate the contribution of mesh-based spatial modeling and continuous-time temporal dynamics. All variants use the same training data, covariates, reconstruction target, and downstream survival-analysis pipeline as the proposed model; hyperparameters are optimized separately for each variant; the latent dimension was fixed at $D=64$.

\begin{itemize}
    \item \textbf{Linear+ODE} keeps the latent ODE dynamics but replaces the graph-based encoder and decoder with linear mappings. This tests whether explicitly modeling mesh connectivity improves reconstruction and downstream prediction.
    \item \textbf{GNN+Fourier} keeps the graph-based mesh encoder and decoder but replaces the latent ODE with a fixed Fourier temporal representation. Subject embeddings are obtained from the corresponding Fourier coefficients, providing a smooth but constrained summary of periodic motion.
    \item \textbf{GNN+Transformer} keeps the graph-based spatial architecture but replaces the latent ODE with a trans\-former-based temporal module that aggregates and expands frame-wise latent states. This tests a discrete sequence model in place of continuous-time dynamics.
\end{itemize}

We also include two published external benchmark architectures, MeshHeart~\citep{qiao2025personalized} and 4D CardioSynth \citep{dou20254d}. For MeshHeart, a published graph-based temporal model for personalized cardiac motion representation, we took the authors' architecture code and retrained it from scratch on the identical mesh data and development split as all other models, using our own training code. Due to GPU memory constraints, batch size was fixed to 2. Using the same early-stopping criterion as for every other model variant, we finetune the learning rate. 
For 4D CardioSynth there is no public code available. We reimplemented the architecture from the paper as faithfully as possible, and tuned its learning rate following the same procedure as for MeshHeart. Several implementation details were left unspecified by the original paper and required judgment calls: we implement the temporal transition as a deterministic residual step (rather than a stochastic one, since no per-step KL term is specified), fix the Chebyshev convolution polynomial order to $K=6$, and use simple addition to compose shape- and motion-decoder outputs. Also, all samples within a training batch are decoded to a single, random timestep, which enables the use of a higher batch size (128) as reported in the paper. As opposed to the paper, we set the latent dimension to 64 (instead of 32) to enable a fair comparison with the other methods.

\section{Heart Rate as an Auxiliary Predictor}
\label{sec:appendix_c}

\begin{figure}
  \centering
  \includegraphics[width=\linewidth]{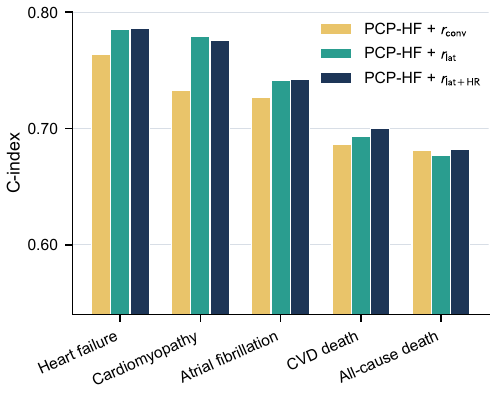}
    \caption{Sex-stratified C-indices for secondary cardiovascular endpoints using the conventional CMR-index score \rconv{}, the latent score \rlat{}, or the latent score augmented with heart rate $r_\mathrm{lat+HR}$ alongside PCP-HF predictors. Aside from mortality prediction, addition of heart rate does not improve over the latent score.}\label{fig:secondary_comparison_supp}
\end{figure}

In the main model, heart rate only aligns cardiac phase through temporal warping; it is not a direct predictor of future events. Because heart rate may carry prognostic information, we tested whether adding it to the latent risk score improves discrimination. Figure~\ref{fig:secondary_comparison_supp} shows only negligible changes and no consistent improvement over the latent score alone.

\section{Supplementary Figures}
\label{sec:appendix_d}

\begin{figure*}
  \centering
  \includegraphics[width=\linewidth]{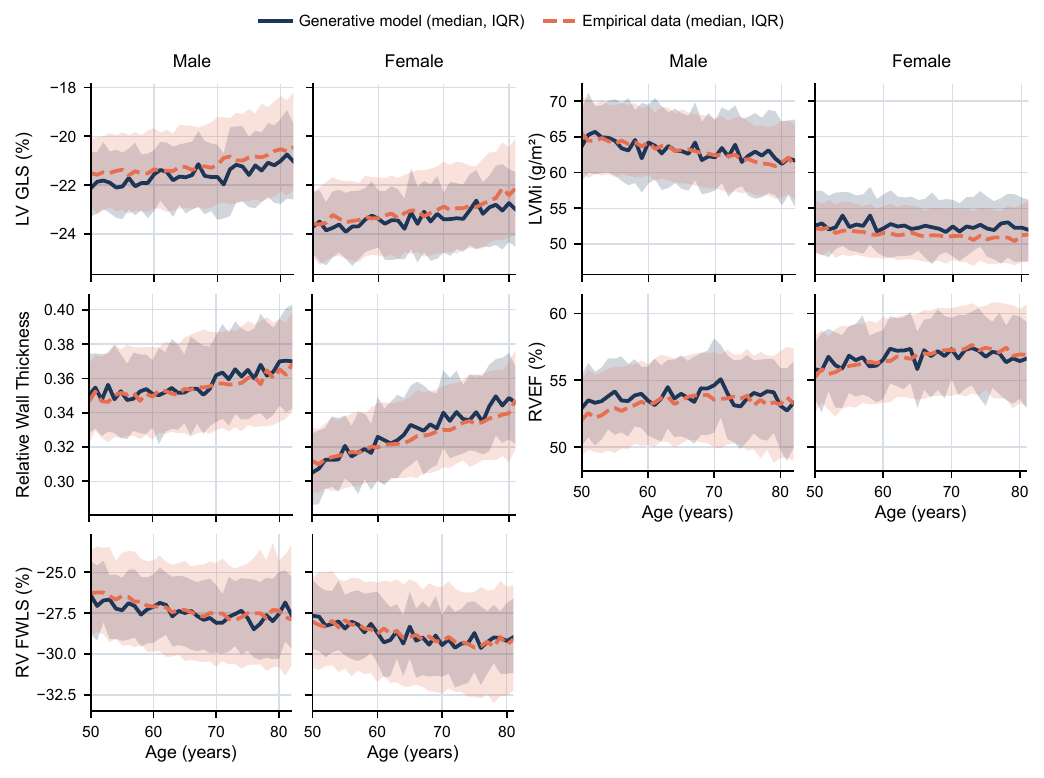}
    \caption{Age- and sex-stratified comparison of LV global longitudinal strain (GLS), LV mass index (LVMi), relative wall thickness, RVEF, and RV free-wall longitudinal strain (FWLS) (from top to bottom) between samples from the generative model (solid blue line, shaded IQR) and the empirical cohort (dashed orange line, shaded IQR). Male subjects are shown on the left and female subjects on the right.}\label{fig:covariate_traversal_other}
\end{figure*}

\begin{figure}
  \centering
  \includegraphics[width=\linewidth]{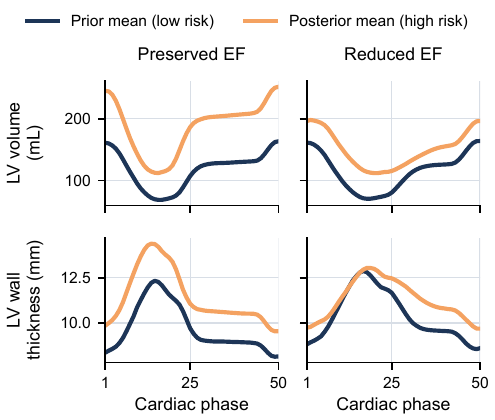}
    \caption{Comparison of the covariate-conditioned prior-mean and posterior-mean reconstructions for two representative high-risk male subjects. Left: preserved ejection fraction. Right: reduced ejection fraction. Top row: LV volume curves. Bottom row: LV wall thickness. Blue denotes the prior mean and orange the posterior mean inferred from the observed mesh sequence.}\label{fig:volume_curves_supp}
\end{figure}

\begin{figure*}
  \centering
  \includegraphics[width=\linewidth]{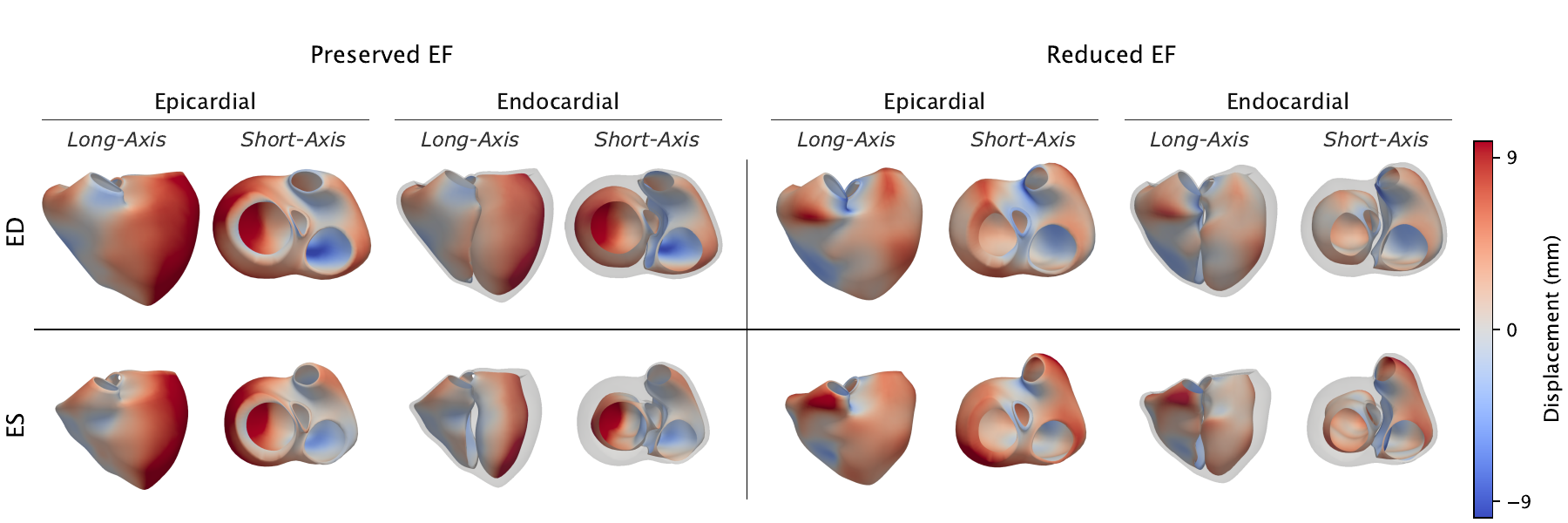}
    \caption{Surface displacement of the mesh decoded from the posterior mean relative to the subject-specific prior mean for the two male subjects in Figure~\ref{fig:volume_curves_supp}. Rows show end diastole (ED) and end systole (ES); within each subject, columns show epicardial and endocardial surfaces rendered in two orthogonal orientations: a long-axis and short-axis view. Red indicates outward displacement and blue inward displacement along the surface normal.}\label{fig:mesh_vis_supp}
\end{figure*}

This section provides supplementary figures. Figure~\ref{fig:covariate_traversal_other} extends Figure~\ref{fig:covariate_traversal_main} to the remaining five CMR markers. Figures~\ref{fig:volume_curves_supp} and \ref{fig:mesh_vis_supp} extend Figures~\ref{fig:volume_curves} and \ref{fig:mesh_vis} with representative high-risk male reconstructions and surface displacements.

\section{Sensitivity to the Cox Penalty}
\label{sec:appendix_e}

\begin{figure}
    \centering
    \includegraphics[width=\linewidth]{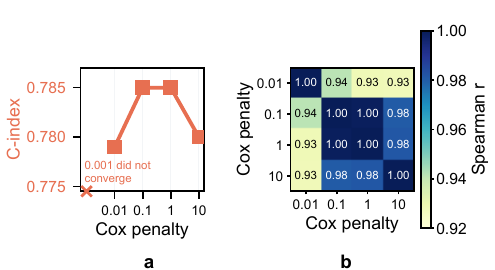}
    \caption{Sensitivity to the Cox penalty weight. Panel \textbf{a} shows C-index as a function of the Cox penalty coefficient; fit did not converge at 0.001. Panel \textbf{b} shows pairwise Spearman correlation between risk scores predicted at each penalty value.}
    \label{fig:sensitivity_supp}
\end{figure}

The penalty parameter of the spline Cox model (Section~\ref{subsec:risk_score}) was selected by 5-fold cross-validation on the development cohort over a logarithmic grid from $10^{-3}$ to $10^{1}$, using held-out partial log-likelihood as the selection criterion. Fitting did not converge at the smallest grid value ($10^{-3}$); Panel~\textbf{a} in Figure~\ref{fig:sensitivity_supp} shows the resulting C-index for the remaining candidate values, which is stable at 0.785 for penalty $0.1$--$1.0$ and only slightly lower for penalties above and below.

We additionally assessed whether this choice of penalty affects subject-level risk ranking. Panel~\textbf{b} in Figure~\ref{fig:sensitivity_supp} reports the pairwise Spearman rank correlation of \rlat{} across the evaluation cohort between refits at different penalty values, which remains high ($r \geq 0.93$) throughout.

\section{Runtime and Memory Comparison}
\label{sec:appendix_f}

All experiments were run on a single NVIDIA RTX 4090 GPU. Table~\ref{tab:runtime_comparison} summarizes training time, inference latency, and peak GPU memory usage during inference for the models in Table~\ref{tab:baseline_comparison}. Inference includes a full run through both encoder and decoder.
MeshHeart's per-frame Chamfer-distance and Laplacian-smoothness losses contribute to its high training memory requirements, restricting the batch size to 2 and increasing both per-epoch and total training time.
In contrast, 4D CardioSynth encodes and decodes only one time step per training iteration, allowing a batch size of 128 and yielding the shortest per-epoch training time. Full-cycle inference is slower, however, because all time steps must be decoded using Chebyshev convolutions, each requiring a polynomial recursion on the graph Laplacian with six sequential sparse matrix multiplications ($K=6$).
The ODE-based models (Linear+ODE and GNN+ODE) have higher inference latency than the feed-forward GNN baselines because of the solver steps required to integrate the latent trajectory. GNN+ODE nevertheless maintains an inference memory footprint comparable to GNN+Fourier and GNN+Transformer, providing continuous-time latent dynamics at the cost of additional runtime rather than higher peak inference memory usage.

\begin{table*}
\caption{Training time, inference latency, and peak inference GPU memory usage for the proposed model and benchmark architectures on a single NVIDIA RTX 4090 GPU.}\label{tab:runtime_comparison}
\centering
\footnotesize
\begin{tabular*}{\tblwidth}{@{} Lccccc@{} }
\toprule
\multirow{2}{*}{Model} & \multicolumn{2}{c}{Train} && \multicolumn{2}{c}{Inference} \\
\cmidrule{2-3} \cmidrule{5-6}
& Time/epoch (min) & Total time (h) && Time/subject (ms) & Peak GPU memory (MB) \\
\midrule
Linear+ODE & 13.0 & 8.7 && 125 & 72 \\
GNN+Fourier & 11.2 & 9.6 && 58 & 1123 \\
GNN+Transformer & 11.6 & 11.8 && 79 & 1132 \\
\midrule
4D CardioSynth & 9.0 & 15.5 && 155 & 1507 \\
MeshHeart & 52.3 & 51.4 && 88 & 3975 \\
\midrule
GNN+ODE & 21.0 & 19.9 && 131 & 1121 \\
\bottomrule
\end{tabular*}
\end{table*}

\section{Posterior Uncertainty in the Latent Risk Score}
\label{sec:appendix_g}

Section~\ref{subsec:robustness} reports the effect of propagating the full posterior over $\mathbf{z}(0)$, rather than only its mean, through to the latent risk score \rlat{}. For each evaluation-cohort subject, we drew $K=50$ samples $\mathbf{z}^{(k)}(0) \sim q(\mathbf{z}(0)\mid\cdot)$ from the already-fitted encoder posterior and scored each sample with the already-fitted Cox model, without retraining. The resulting sample-averaged C-index (0.779) is close to, but slightly below, the point-estimate C-index (0.785); because \rlat{} is a nonlinear (spline-transformed) function of $\mathbf{z}(0)$, averaging predictions over noisy samples is not equivalent to evaluating the function at the mean.

\begin{figure}
    \centering
    \includegraphics[width=\linewidth]{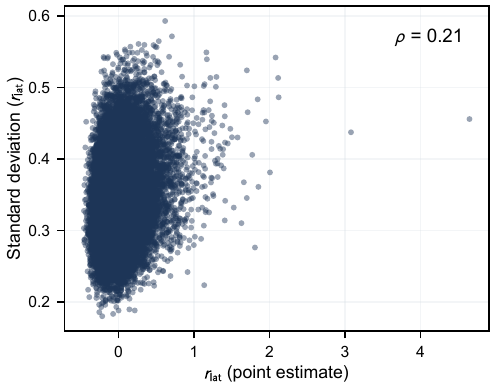}
    \caption{Per-subject posterior standard deviation of \rlat{} (from $K=50$ posterior samples) versus the point-estimate \rlat{}.}
    \label{fig:sd_r_lat_vs_r_lat}
\end{figure}

Figure~\ref{fig:sd_r_lat_vs_r_lat} plots the per-subject posterior standard deviation of \rlat{} against the point-estimate \rlat{}. Uncertainty increases with predicted risk (Spearman $r=0.206$), while it shows no comparable association with age ($r=0.023$) or with $\|\hat{\mathbf{z}}\|_2$, the magnitude of the (direction-agnostic) normalized deviation from the covariate-conditioned prior ($r=0.025$). The current risk score does not communicate this heterogeneity: two subjects with the same point-estimate \rlat{} may differ considerably in how confidently that estimate is held.

\printcredits

\section*{Acknowledgments}

This research has been conducted using the UK Biobank Resources under application numbers 81959 and 206575.
The project was funded by Dataspectrum4CVD from the Swiss Data Science Center \#2022-812/Personalized Health \& Related Technologies C22-15P, Zurich, Switzerland. OVD reports funding from the NIH NHLBI (K01 HL135342, R21 HL167173) and the American Heart Association (17IG\allowbreak{}MV33860009, 26SFRNCVHD1592182, 26SFRNPVHD162\allowbreak{}3048).

\section*{Declaration of AI-assisted technologies in the manuscript preparation process}

During the preparation of this work, the authors used Claude by Anthropic to improve the readability and language of the manuscript. After using this tool, the authors reviewed and edited the content as needed and take full responsibility for the content of the published article.

\bibliographystyle{cas-model2-names}

\bibliography{cas-refs}

@CONFERENCE{kingma2013auto,
  title={Auto-encoding variational bayes},
  author={Kingma, Diederik P and Welling, Max},
  booktitle={International conference on learning representations},
  pages={0--0},
  year={2014}
}

@CONFERENCE{khemakhem2020variational,
  title={Variational autoencoders and nonlinear ica: A unifying framework},
  author={Khemakhem, Ilyes and Kingma, Diederik and Monti, Ricardo and Hyvarinen, Aapo},
  booktitle={International conference on artificial intelligence and statistics},
  pages={2207--2217},
  year={2020},
}

@CONFERENCE{gong2019spiralnet++,
  title={Spiralnet++: A fast and highly efficient mesh convolution operator},
  author={Gong, Shunwang and Chen, Lei and Bronstein, Michael and Zafeiriou, Stefanos},
  booktitle={Proceedings of the IEEE/CVF international conference on computer vision workshops},
  pages={0--0},
  year={2019}
}

@CONFERENCE{sohn2015learning,
  title={Learning structured output representation using deep conditional generative models},
  author={Sohn, Kihyuk and Lee, Honglak and Yan, Xinchen},
  booktitle={Advances in neural information processing systems},
  pages={0--0},
  year={2015}
}

@ARTICLE{qiao2025personalized,
  title={A personalized time-resolved 3D mesh generative model for unveiling normal heart dynamics},
  author={Qiao, Mengyun and McGurk, Kathryn A and Wang, Shuo and Matthews, Paul M and O’Regan, Declan P and Bai, Wenjia},
  journal={Nature Machine Intelligence},
  volume={7},
  pages={800--811},
  year={2025},
  doi={10.1038/s42256-025-01035-5},
}

@ARTICLE{khan201910,
  title={10-year risk equations for incident heart failure in the general population},
  author={Khan, Sadiya S and Ning, Hongyan and Shah, Sanjiv J and Yancy, Clyde W and Carnethon, Mercedes and Berry, Jarett D and Mentz, Robert J and O’Brien, Emily and Correa, Adolfo and Suthahar, Navin and others},
  journal={Journal of the American College of Cardiology},
  volume={73},
  pages={2388--2397},
  year={2019},
  doi={10.1016/j.jacc.2019.02.057},
}

@ARTICLE{elliott2020predictive,
  title={Predictive accuracy of a polygenic risk score--enhanced prediction model vs a clinical risk score for coronary artery disease},
  author={Elliott, Joshua and Bodinier, Barbara and Bond, Tom A and Chadeau-Hyam, Marc and Evangelou, Evangelos and Moons, Karel GM and Dehghan, Abbas and Muller, David C and Elliott, Paul and Tzoulaki, Ioanna},
  journal={Jama},
  volume={323},
  pages={636--645},
  year={2020},
  doi={10.1001/jama.2019.22241},
}

@ARTICLE{klarin2019genome,
  title={Genome-wide association study of peripheral artery disease in the Million Veteran Program},
  author={Klarin, Derek and Lynch, Julie and Aragam, Krishna and Chaffin, Mark and Assimes, Themistocles L and Huang, Jie and Lee, Kyung Min and Shao, Qing and Huffman, Jennifer E and Natarajan, Pradeep and others},
  journal={Nature medicine},
  volume={25},
  pages={1274--1279},
  year={2019},
  doi={10.1038/s41591-019-0492-5},
}

@CONFERENCE{dillon2025open,
  title={An open-source end-to-end pipeline for generating 3D+t biventricular meshes from cardiac magnetic resonance imaging},
  author={Dillon, Joshua R and Mauger, Charl{\`e}ne and Zhao, Debbie and Deng, Yu and Petersen, Steffen E and McCulloch, Andrew D and Young, Alistair A and Nash, Martyn P},
  booktitle={International Conference on Functional Imaging and Modeling of the Heart},
  pages={372--383},
  year={2025},
}

@ARTICLE{du1916clinical,
  title={A formula to estimate the approximate surface area if height and weight be known},
  author={Du Bois, Delafield and Du Bois, Eugene F},
  journal={Archives of internal medicine},
  volume={17},
  pages={863--871},
  year={1916},
  doi={10.1001/archinte.1916.00080130010002},
}

@ARTICLE{van2011mice,
  title={mice: Multivariate imputation by chained equations in R},
  author={Van Buuren, Stef and Groothuis-Oudshoorn, Karin},
  journal={Journal of statistical software},
  volume={45},
  pages={1--67},
  year={2011},
  doi={10.18637/jss.v045.i03},
}

@CONFERENCE{chen2018neural,
  title={Neural ordinary differential equations},
  author={Chen, Ricky TQ and Rubanova, Yulia and Bettencourt, Jesse and Duvenaud, David K},
  booktitle={Advances in neural information processing systems},
  pages={0--0},
  year={2018}
}

@ARTICLE{piras2017morphologically,
  title={Morphologically normalized left ventricular motion indicators from MRI feature tracking characterize myocardial infarction},
  author={Piras, Paolo and Teresi, Luciano and Puddu, Paolo Emilio and Torromeo, Concetta and Young, Alistair A and Suinesiaputra, Avan and Medrano-Gracia, Pau},
  journal={Scientific reports},
  volume={7},
  pages={12259},
  year={2017},
  doi={10.1038/s41598-017-12539-5},
}

@ARTICLE{gilbert2019independent,
  title={Independent left ventricular morphometric atlases show consistent relationships with cardiovascular risk factors: a UK biobank study},
  author={Gilbert, Kathleen and Bai, Wenjia and Mauger, Charlene and Medrano-Gracia, Pau and Suinesiaputra, Avan and Lee, Aaron M and Sanghvi, Mihir M and Aung, Nay and Piechnik, Stefan K and Neubauer, Stefan and others},
  journal={Scientific reports},
  volume={9},
  pages={1130},
  year={2019},
  doi={10.1038/s41598-018-37916-6},
}

@ARTICLE{mauger2022multi,
  title={Multi-ethnic study of atherosclerosis: relationship between left ventricular shape at cardiac MRI and 10-year outcomes},
  author={Mauger, Charl{\`e}ne A and Gilbert, Kathleen and Suinesiaputra, Avan and Bluemke, David A and Wu, Colin O and Lima, Jo{\~a}o AC and Young, Alistair A and Ambale-Venkatesh, Bharath},
  journal={Radiology},
  volume={306},
  pages={e220122},
  year={2022},
  doi={10.1148/radiol.220122},
}

@CONFERENCE{dou2023conditional,
  title={A conditional flow variational autoencoder for controllable synthesis of virtual populations of anatomy},
  author={Dou, Haoran and Ravikumar, Nishant and Frangi, Alejandro F},
  booktitle={International Conference on Medical Image Computing and Computer-Assisted Intervention},
  pages={143--152},
  year={2023},
}

@ARTICLE{beetz2022interpretable,
  title={Interpretable cardiac anatomy modeling using variational mesh autoencoders},
  author={Beetz, Marcel and Corral Acero, Jorge and Banerjee, Abhirup and Eitel, Ingo and Zacur, Ernesto and Lange, Torben and Stiermaier, Thomas and Evertz, Ruben and Backhaus, S{\"o}ren J and Thiele, Holger and others},
  journal={Frontiers in cardiovascular medicine},
  volume={9},
  pages={983868},
  year={2022},
  doi={10.3389/fcvm.2022.983868},
}

@CONFERENCE{biffi2018learning,
  title={Learning interpretable anatomical features through deep generative models: Application to cardiac remodeling},
  author={Biffi, Carlo and Oktay, Ozan and Tarroni, Giacomo and Bai, Wenjia and De Marvao, Antonio and Doumou, Georgia and Rajchl, Martin and Bedair, Reem and Prasad, Sanjay and Cook, Stuart and others},
  booktitle={International conference on medical image computing and computer-assisted intervention},
  pages={464--471},
  year={2018},
}

@ARTICLE{marquand2016understanding,
  title={Understanding heterogeneity in clinical cohorts using normative models: beyond case-control studies},
  author={Marquand, Andre F and Rezek, Iead and Buitelaar, Jan and Beckmann, Christian F},
  journal={Biological Psychiatry},
  volume={80},
  number={7},
  pages={552--561},
  year={2016},
  doi={10.1016/j.biopsych.2015.12.023},
}

@ARTICLE{bello2019deep,
  title={Deep-learning cardiac motion analysis for human survival prediction},
  author={Bello, Ghalib A and Dawes, Timothy JW and Duan, Jinming and Biffi, Carlo and De Marvao, Antonio and Howard, Luke SGE and Gibbs, J Simon R and Wilkins, Martin R and Cook, Stuart A and Rueckert, Daniel and others},
  journal={Nature machine intelligence},
  volume={1},
  pages={95--104},
  year={2019},
  doi={10.1038/s42256-019-0019-2},
}

@CONFERENCE{vaswani2017attention,
  title={Attention is all you need},
  author={Vaswani, Ashish and Shazeer, Noam and Parmar, Niki and Uszkoreit, Jakob and Jones, Llion and Gomez, Aidan N and Kaiser, {\L}ukasz and Polosukhin, Illia},
  booktitle={Advances in neural information processing systems},
  pages={0--0},
  year={2017}
}

@ARTICLE{chung2004duration,
  title={Duration of diastole and its phases as a function of heart rate during supine bicycle exercise},
  author={Chung, Charles S and Karamanoglu, Mustafa and Kov{\'a}cs, S{\'a}ndor J},
  journal={American Journal of Physiology-Heart and Circulatory Physiology},
  volume={287},
  pages={H2003--H2008},
  year={2004},
  doi={10.1152/ajpheart.00404.2004},
}

@ARTICLE{petersen2016uk,
  title={UK Biobank's cardiovascular magnetic resonance protocol},
  author={Petersen, Steffen E and Matthews, Paul M and Francis, Jane M and Robson, Matthew D and Zemrak, Filip and Boubertakh, Redha and Young, Alistair A and Hudson, Sarah and Weale, Peter and Garratt, Steve and others},
  journal={Journal of cardiovascular magnetic resonance},
  volume={18},
  pages={8},
  year={2016},
  doi={10.1186/s12968-016-0227-4},
}

@ARTICLE{meunier2021202,
  title={2021 ESC guidelines for the diagnosis and treatment of acute and chronic heart failure},
  author={Mcdonagh, Theresa A and Metra, Marco and Adamo, Marianna and Gardner, Roy S and Baumbach, Andreas and B{\"o}hm, Michael and Burri, Haran and Butler, Javed and {\v{C}}elutkien{\.e}, Jelena and Chioncel, Ovidiu and others},
  journal={European Heart Journal},
  volume={42},
  pages={3599--3726},
  year={2021},
  doi={10.1093/eurheartj/ehab368},
}

@ARTICLE{heidenreich20222022,
  title={2022 AHA/ACC/HFSA guideline for the management of heart failure: a report of the American College of Cardiology/American Heart Association Joint Committee on Clinical Practice Guidelines},
  author={Heidenreich, Paul A and Bozkurt, Biykem and Aguilar, David and Allen, Larry A and Byun, Joni J and Colvin, Monica M and Deswal, Anita and Drazner, Mark H and Dunlay, Shannon M and Evers, Linda R and others},
  journal={Journal of the American College of Cardiology},
  volume={79},
  pages={e263--e421},
  year={2022},
  doi={10.1161/CIR.0000000000001063},
}

@ARTICLE{bronstein2017geometric,
  title={Geometric deep learning: going beyond euclidean data},
  author={Bronstein, Michael M and Bruna, Joan and LeCun, Yann and Szlam, Arthur and Vandergheynst, Pierre},
  journal={IEEE Signal Processing Magazine},
  volume={34},
  pages={18--42},
  year={2017},
  doi={10.1109/MSP.2017.2693418},
}

@ARTICLE{fonseca2011cardiac,
  title={The Cardiac Atlas Project—an imaging database for computational modeling and statistical atlases of the heart},
  author={Fonseca, Carissa G and Backhaus, Michael and Bluemke, David A and Britten, Randall D and Chung, Jae Do and Cowan, Brett R and Dinov, Ivo D and Finn, J Paul and Hunter, Peter J and Kadish, Alan H and others},
  journal={Bioinformatics},
  volume={27},
  pages={2288--2295},
  year={2011},
  doi={10.1093/bioinformatics/btr360},
}

@ARTICLE{bijnens2012myocardial,
  title={Myocardial motion and deformation: What does it tell us and how does it relate to function?},
  author={Bijnens, Bart and Cikes, M and Butakoff, C and Sitges, Marta and Crispi, Fatima},
  journal={Fetal diagnosis and therapy},
  volume={32},
  pages={5--16},
  year={2012},
  doi={10.1159/000335649},
}

@CONFERENCE{mauger2018iterative,
  title={An iterative diffeomorphic algorithm for registration of subdivision surfaces: application to congenital heart disease},
  author={Mauger, Charlene and Gilbert, Kathleen and Suinesiaputra, Avan and Pontre, Beau and Omens, J and McCulloch, A and Young, A},
  booktitle={2018 40th Annual International Conference of the IEEE Engineering in Medicine and Biology Society (EMBC)},
  pages={596--599},
  year={2018},
}

@CONFERENCE{kanazawa2019learning,
  title={Learning 3d human dynamics from video},
  author={Kanazawa, Angjoo and Zhang, Jason Y and Felsen, Panna and Malik, Jitendra},
  booktitle={Proceedings of the IEEE/CVF conference on computer vision and pattern recognition},
  pages={5614--5623},
  year={2019}
}

@CONFERENCE{achlioptas2018learning,
  title={Learning representations and generative models for 3d point clouds},
  author={Achlioptas, Panos and Diamanti, Olga and Mitliagkas, Ioannis and Guibas, Leonidas},
  booktitle={International conference on machine learning},
  pages={40--49},
  year={2018},
}

@MISC{who_cvd_global_2025,
  author       = {{World Health Organization}},
  title        = {Cardiovascular diseases (CVDs)},
  year         = {2025},
  url          = {https://www.who.int/health-topics/cardiovascular-diseases},
  note         = {Accessed 2026-04-22}
}

@ARTICLE{kriegeskorte2008representational,
  title={Representational similarity analysis-connecting the branches of systems neuroscience},
  author={Kriegeskorte, Nikolaus and Mur, Marieke and Bandettini, Peter A},
  journal={Frontiers in systems neuroscience},
  volume={2},
  pages={249},
  year={2008},
  doi={10.3389/neuro.06.004.2008},
}

@CONFERENCE{paszke2019pytorch,
  title={Pytorch: An imperative style, high-performance deep learning library},
  author={Paszke, Adam and Gross, Sam and Massa, Francisco and Lerer, Adam and Bradbury, James and Chanan, Gregory and Killeen, Trevor and Lin, Zeming and Gimelshein, Natalia and Antiga, Luca and others},
  booktitle={Advances in neural information processing systems},
  pages={0--0},
  year={2019}
}

@ARTICLE{dormand1980family,
  title={A family of embedded Runge-Kutta formulae},
  author={Dormand, John R and Prince, Peter J},
  journal={Journal of computational and applied mathematics},
  volume={6},
  pages={19--26},
  year={1980},
  doi={10.1016/0771-050X(80)90013-3},
}

@CONFERENCE{kingma2014adam,
  title={Adam: A method for stochastic optimization},
  author={Kingma, Diederik P and Ba, Jimmy},
  booktitle={International conference for learning representations},
  pages={0--0},
  year={2015}
}

@ARTICLE{DavidsonPilon2019,
  year = {2019},
  volume = {4},
  pages = {1317},
  author = {Cameron Davidson-Pilon},
  title = {lifelines: survival analysis in Python},
  journal = {Journal of Open Source Software},
  doi={10.21105/joss.01317},
}

@ARTICLE{fry2017comparison,
  title={Comparison of sociodemographic and health-related characteristics of UK Biobank participants with those of the general population},
  author={Fry, Anna and Littlejohns, Thomas J and Sudlow, Cathie and Doherty, Nicola and Adamska, Ligia and Sprosen, Tim and Collins, Rory and Allen, Naomi E},
  journal={American journal of epidemiology},
  volume={186},
  pages={1026--1034},
  year={2017},
  doi={10.1093/aje/kwx246},
}

@ARTICLE{aziz2013diastolic,
  title={Diastolic heart failure: a concise review},
  author={Aziz, Fahad and Luqman-Arafath, TK and Enweluzo, Chijioke and Dutta, Simanta and Zaeem, Misbah},
  journal={Journal of clinical medicine research},
  volume={5},
  pages={327},
  year={2013},
  doi={10.4021/jocmr1532w},
}

@CONFERENCE{sinclair2017fully,
  title={Fully automated segmentation-based respiratory motion correction of multiplanar cardiac magnetic resonance images for large-scale datasets},
  author={Sinclair, Matthew and Bai, Wenjia and Puyol-Ant{\'o}n, Esther and Oktay, Ozan and Rueckert, Daniel and King, Andrew P},
  booktitle={International Conference on Medical Image Computing and Computer-Assisted Intervention},
  pages={332--340},
  year={2017},
}

@ARTICLE{isensee2021nnu,
  title={nnU-Net: a self-configuring method for deep learning-based biomedical image segmentation},
  author={Isensee, Fabian and Jaeger, Paul F and Kohl, Simon AA and Petersen, Jens and Maier-Hein, Klaus H},
  journal={Nature methods},
  volume={18},
  pages={203--211},
  year={2021},
  doi={10.1038/s41592-020-01008-z},
}

@ARTICLE{pennell2010cardiovascular,
  title={Cardiovascular magnetic resonance},
  author={Pennell, Dudley J},
  journal={Circulation},
  volume={121},
  pages={692--705},
  year={2010},
  doi={10.1161/CIRCULATIONAHA.108.811547},
}

@ARTICLE{littlejohns2020ukbiobank,
  title={The UK Biobank imaging enhancement of 100,000 participants: rationale, data collection, management and future directions},
  author={Littlejohns, Thomas J and Holliday, Jo and Gibson, Lorna M and Garratt, Steve and Oesingmann, Niels and Alfaro-Almagro, Fidel and Bell, Jimmy D and Boultwood, Chris and Collins, Rory and Conroy, Megan C and others},
  journal={Nature communications},
  volume={11},
  pages={2624},
  year={2020},
  doi={10.1038/s41467-020-15948-9},
}

@ARTICLE{zhang2010cardiac4d,
  title={4-D cardiac MR image analysis: left and right ventricular morphology and function},
  author={Zhang, Honghai and Wahle, Andreas and Johnson, Ryan K and Scholz, Thomas D and Sonka, Milan},
  journal={IEEE Transactions on Medical Imaging},
  volume={29},
  number={2},
  pages={350--364},
  year={2010},
  doi={10.1109/TMI.2009.2030799},
}

@ARTICLE{liu2017spatiotemporal,
  title={Spatiotemporal Strategies for Joint Segmentation and Motion Tracking From Cardiac Image Sequences},
  author={Liu, Huafeng and Wang, Ting and Xu, Lei and Shi, Pengcheng},
  journal={IEEE Journal of Translational Engineering in Health and Medicine},
  volume={5},
  pages={1800219},
  year={2017},
  doi={10.1109/JTEHM.2017.2665496}
}

@ARTICLE{guo2022uncertaintyshape,
  title={Cardiac MRI segmentation with sparse annotations: Ensembling deep learning uncertainty and shape priors},
  author={Guo, Fumin and Ng, Matthew and Kuling, Grey and Wright, Graham},
  journal={Medical Image Analysis},
  volume={81},
  pages={102532},
  year={2022},
  doi={10.1016/j.media.2022.102532}
}

@ARTICLE{ng2023uncertainty,
  title={Estimating Uncertainty in Neural Networks for Cardiac MRI Segmentation: A Benchmark Study},
  author={Ng, Matthew and Guo, Fumin and Biswas, Labonny and Petersen, Steffen E and Piechnik, Stefan K and Neubauer, Stefan and Wright, Graham},
  journal={IEEE Transactions on Biomedical Engineering},
  volume={70},
  number={6},
  pages={1955--1966},
  year={2023},
  doi={10.1109/TBME.2022.3232730}
}

@ARTICLE{meng2024deepmesh,
  title={DeepMesh: Mesh-Based Cardiac Motion Tracking Using Deep Learning},
  author={Meng, Qingjie and Bai, Wenjia and O'Regan, Declan P and Rueckert, Daniel},
  journal={IEEE Transactions on Medical Imaging},
  volume={43},
  number={4},
  pages={1489--1500},
  year={2024},
  doi={10.1109/TMI.2023.3340118}
}

@ARTICLE{kong2021wholeheart,
  title={A deep-learning approach for direct whole-heart mesh reconstruction},
  author={Kong, Fanwei and Wilson, Nathan and Shadden, Shawn},
  journal={Medical Image Analysis},
  volume={74},
  pages={102222},
  year={2021},
  doi={10.1016/j.media.2021.102222}
}

@ARTICLE{corral2022understanding,
  title={Understanding and improving risk assessment after myocardial infarction using automated left ventricular shape analysis},
  author={Corral Acero, Jorge and Schuster, Andreas and Zacur, Ernesto and Lange, Torben and Stiermaier, Thomas and Backhaus, S{\"o}ren J and Thiele, Holger and Bueno-Orovio, Alfonso and Lamata, Pablo and Eitel, Ingo and others},
  journal={Cardiovascular Imaging},
  volume={15},
  pages={1563--1574},
  year={2022},
  doi={10.1016/j.jcmg.2021.11.027},
}

@ARTICLE{bai2015bi,
  title={A bi-ventricular cardiac atlas built from 1000+ high resolution MR images of healthy subjects and an analysis of shape and motion},
  author={Bai, Wenjia and Shi, Wenzhe and de Marvao, Antonio and Dawes, Timothy JW and O’Regan, Declan P and Cook, Stuart A and Rueckert, Daniel},
  journal={Medical image analysis},
  volume={26},
  pages={133--145},
  year={2015},
  doi={10.1016/j.media.2015.08.009},
}

@ARTICLE{xia2022automatic,
  title={Automatic 3D+t four-chamber CMR quantification of the UK biobank: integrating imaging and non-imaging data priors at scale},
  author={Xia, Yan and Chen, Xiang and Ravikumar, Nishant and Kelly, Christopher and Attar, Rahman and Aung, Nay and Neubauer, Stefan and Petersen, Steffen E and Frangi, Alejandro F},
  journal={Medical Image Analysis},
  volume={80},
  pages={102498},
  year={2022},
  doi={10.1016/j.media.2022.102498},
}

@ARTICLE{frangi2003automatic,
  title={Automatic construction of multiple-object three-dimensional statistical shape models: Application to cardiac modeling},
  author={Frangi, Alejandro F and Rueckert, Daniel and Schnabel, Julia A and Niessen, Wiro J},
  journal={IEEE transactions on medical imaging},
  volume={21},
  pages={1151--1166},
  year={2003},
  doi={10.1109/TMI.2002.804426},
}

@ARTICLE{suinesiaputra2017statistical,
  title={Statistical shape modeling of the left ventricle: myocardial infarct classification challenge},
  author={Suinesiaputra, Avan and Ablin, Pierre and Alba, Xenia and Alessandrini, Martino and Allen, Jack and Bai, Wenjia and Cimen, Serkan and Claes, Peter and Cowan, Brett R and D’hooge, Jan and others},
  journal={IEEE journal of biomedical and health informatics},
  volume={22},
  pages={503--515},
  year={2017},
  doi={10.1109/JBHI.2017.2652449},
}

@ARTICLE{gaggion2025multi,
  title={Multi-view hybrid graph convolutional network for volume-to-mesh reconstruction in cardiovascular MRI},
  author={Gaggion, Nicol{\'a}s and Matheson, Benjamin A and Xia, Yan and Bonazzola, Rodrigo and Ravikumar, Nishant and Taylor, Zeike A and Milone, Diego H and Frangi, Alejandro F and Ferrante, Enzo},
  journal={Medical Image Analysis},
  volume={104},
  pages={103630},
  year={2025},
  doi={10.1016/j.media.2025.103630},
}

@ARTICLE{kong2023learning,
  title={Learning whole heart mesh generation from patient images for computational simulations},
  author={Kong, Fanwei and Shadden, Shawn C},
  journal={IEEE Transactions on Medical Imaging},
  volume={42},
  pages={533--545},
  year={2023},
  doi={10.1109/TMI.2022.3219284},
}

@ARTICLE{dawes2017machine,
  title={Machine learning of three-dimensional right ventricular motion enables outcome prediction in pulmonary hypertension: a cardiac MR imaging study},
  author={Dawes, Timothy JW and de Marvao, Antonio and Shi, Wenzhe and Fletcher, Tristan and Watson, Geoffrey MJ and Wharton, John and Rhodes, Christopher J and Howard, Luke SGE and Gibbs, J Simon R and Rueckert, Daniel and others},
  journal={Radiology},
  volume={283},
  pages={381--390},
  year={2017},
  doi={10.1148/radiol.2016161315},
}

@ARTICLE{guo2023survival,
  title={Survival prediction of heart failure patients using motion-based analysis method},
  author={Guo, Saidi and Zhang, Heye and Gao, Yifeng and Wang, Hui and Xu, Lei and Gao, Zhifan and Guzzo, Antonella and Fortino, Giancarlo},
  journal={Computer Methods and Programs in Biomedicine},
  volume={236},
  pages={107547},
  year={2023},
  doi={10.1016/j.cmpb.2023.107547},
}

@CONFERENCE{yildiz2019ode2vae,
  title={Ode2vae: Deep generative second order odes with bayesian neural networks},
  author={Yildiz, Cagatay and Heinonen, Markus and Lahdesmaki, Harri},
  booktitle={Advances in Neural Information Processing Systems},
  pages={0--0},
  year={2019}
}

@CONFERENCE{greydanus2019hamiltonian,
  title={Hamiltonian neural networks},
  author={Greydanus, Samuel and Dzamba, Misko and Yosinski, Jason},
  booktitle={Advances in neural information processing systems},
  pages={0--0},
  year={2019}
}

@ARTICLE{wang2009modelling,
  title={Modelling passive diastolic mechanics with quantitative MRI of cardiac structure and function},
  author={Wang, Vicky Y and Lam, Hoi Ieng and Ennis, Daniel B and Cowan, Brett R and Young, Alistair A and Nash, Martyn P},
  journal={Medical image analysis},
  volume={13},
  pages={773--784},
  year={2009},
  doi={10.1016/j.media.2009.07.006},
}

@ARTICLE{muffoletto2024evaluation,
  title={Evaluation of deep learning estimation of whole heart anatomy from automated cardiovascular magnetic resonance short-and long-axis analyses in UK Biobank},
  author={Muffoletto, Marica and Xu, Hao and Burns, Richard and Suinesiaputra, Avan and Nasopoulou, Anastasia and Kunze, Karl P and Neji, Radhouene and Petersen, Steffen E and Niederer, Steven A and Rueckert, Daniel and others},
  journal={European Heart Journal-Cardiovascular Imaging},
  volume={25},
  pages={1374--1383},
  year={2024},
  doi={10.1093/ehjci/jeae123},
}

@ARTICLE{beg2005computing,
  title={Computing large deformation metric mappings via geodesic flows of diffeomorphisms},
  author={Beg, M Faisal and Miller, Michael I and Trouv{\'e}, Alain and Younes, Laurent},
  journal={International journal of computer vision},
  volume={61},
  pages={139--157},
  year={2005},
  doi={10.1023/B:VISI.0000043755.93987.aa},
}

@ARTICLE{krebs2019learning,
  title={Learning a probabilistic model for diffeomorphic registration},
  author={Krebs, Julian and Delingette, Herv{\'e} and Mailh{\'e}, Boris and Ayache, Nicholas and Mansi, Tommaso},
  journal={IEEE transactions on medical imaging},
  volume={38},
  pages={2165--2176},
  year={2019},
  doi={10.1109/TMI.2019.2897112},
}

@ARTICLE{dalca2019unsupervised,
  title={Unsupervised learning of probabilistic diffeomorphic registration for images and surfaces},
  author={Dalca, Adrian V and Balakrishnan, Guha and Guttag, John and Sabuncu, Mert R},
  journal={Medical image analysis},
  volume={57},
  pages={226--236},
  year={2019},
  doi={10.1016/j.media.2019.07.006},
}

@CONFERENCE{wu2022nodeo,
  title={Nodeo: A neural ordinary differential equation based optimization framework for deformable image registration},
  author={Wu, Yifan and Jiahao, Tom Z and Wang, Jiancong and Yushkevich, Paul A and Hsieh, M Ani and Gee, James C},
  booktitle={Proceedings of the IEEE/CVF conference on computer vision and pattern recognition},
  pages={20804--20813},
  year={2022}
}

@ARTICLE{ji2022sex,
  title={Sex differences in myocardial and vascular aging},
  author={Ji, Hongwei and Kwan, Alan C and Chen, Melanie T and Ouyang, David and Ebinger, Joseph E and Bell, Susan P and Niiranen, Teemu J and Bello, Natalie A and Cheng, Susan},
  journal={Circulation research},
  volume={130},
  pages={566--577},
  year={2022},
  doi={10.1161/CIRCRESAHA.121.319902},
}

@ARTICLE{chadalavada2025mri,
  title={MRI-derived Right Ventricular Global Longitudinal Strain Predicts Heart Failure},
  author={Chadalavada, Sucharitha and Mahmood, Adil and Salatzki, Janek and Hesse, Kerrick and Fung, Kenneth and Khanji, Mohammed Y and Raisi-Estabragh, Zahra and Aung, Nay and Petersen, Steffen E},
  journal={Radiology: Cardiothoracic Imaging},
  volume={7},
  pages={e240493},
  year={2025},
  doi={10.1148/ryct.240493},
}

@ARTICLE{mercadier2025printmesh,
  title={PrIntMesh: Precise Intersection Surfaces for 3D Organ Mesh Reconstruction},
  author={Mercadier, Deniz Sayin and Le, Hieu and Chen, Yihong and Yang, Jiancheng and Wickramasinghe, Udaranga and Fua, Pascal},
  journal={arXiv preprint arXiv:2511.16186},
  year={2025}
}

@ARTICLE{harrell1982evaluating,
  title={Evaluating the yield of medical tests},
  author={Harrell, Frank E and Califf, Robert M and Pryor, David B and Lee, Kerry L and Rosati, Robert A},
  journal={Jama},
  volume={247},
  number={18},
  pages={2543--2546},
  year={1982},
  doi={doi:10.1001/jama.1982.03320430047030}
}

@ARTICLE{mauger2019right,
  title={Right ventricular shape and function: cardiovascular magnetic resonance reference morphology and biventricular risk factor morphometrics in UK Biobank},
  author={Mauger, Charl{\`e}ne and Gilbert, Kathleen and Lee, Aaron M and Sanghvi, Mihir M and Aung, Nay and Fung, Kenneth and Carapella, Valentina and Piechnik, Stefan K and Neubauer, Stefan and Petersen, Steffen E and others},
  journal={Journal of Cardiovascular Magnetic Resonance},
  volume={21},
  pages={41},
  year={2019},
  doi={10.1186/s12968-019-0551-6}
}

@INPROCEEDINGS{dou20254d,
  title={4d cardiosynth: Synthesising dynamic virtual heart populations through spatiotemporal disentanglement},
  author={Dou, Haoran and Huang, Jinghan and Zakeri, Arezoo and Zhou, Zherui and Mu, Tingting and Duan, Jinming and Frangi, Alejandro F},
  booktitle={International Conference on Medical Image Computing and Computer-Assisted Intervention},
  pages={3--12},
  year={2025},
  organization={Springer}
}

\end{document}